\documentclass[runningheads]{llncs}
\usepackage{graphicx}
\usepackage{comment}

\usepackage{amsmath,amssymb} 
\usepackage{color}

\usepackage[dvipsnames]{xcolor}

\graphicspath{{images/}}
\usepackage{breqn}
\usepackage{multirow}
\usepackage{hhline}
\usepackage{ltablex}
\usepackage{siunitx}
\usepackage{booktabs}
\usepackage{algorithm}
\usepackage{algcompatible}
\usepackage[noend]{algpseudocode}

\usepackage{soul}
\DeclareMathOperator*{\argminA}{\,min} 
\usepackage{mathtools}
\usepackage{arydshln}

\usepackage{capt-of}
\usepackage{float}
\usepackage{wrapfig}
\usepackage{bm}
\usepackage{afterpage}
\usepackage[export]{adjustbox}
\usepackage{amsthm,enumitem}
\def\mathbi#1{\textbf{\em #1}}
\usepackage{tikz,xcolor}
\usepackage[colorlinks = true,
            linkcolor = blue,
            urlcolor  = blue,
            citecolor = green,
            anchorcolor = green]{hyperref}
            
\definecolor{lime}{HTML}{A6CE39}
\DeclareRobustCommand{\orcidicon}{%
	\begin{tikzpicture}
	\draw[lime, fill=lime] (0,0) 
	circle [radius=0.16] 
	node[white] {{\fontfamily{qag}\selectfont \tiny ID}};
	\draw[white, fill=white] (-0.0625,0.095) 
	circle [radius=0.007];
	\end{tikzpicture}
	\hspace{-2mm}}

\foreach \x in {A, ..., Z}{%
	\expandafter\xdef\csname orcid\x\endcsname{\noexpand\href{https://orcid.org/\csname orcidauthor\x\endcsname}{\noexpand\orcidicon}}
}

\begin{document}
\title{Talking-head Generation with Rhythmic Head Motion}

\titlerunning{Talking-head Generation with Rhythmic Head Motion}


\author{Lele Chen$^\dagger$ \orcidA{} \and Guofeng Cui$^\dagger$ \orcidB{} \and Celong Liu$^\ddag$ \orcidH{}\and Zhong Li$^\ddag$\orcidD{} \and Ziyi Kou$^\dagger$ \orcidE{} \and Yi Xu$^\ddag$\orcidF{} \and Chenliang Xu$^\dagger$\orcidG{}}

\authorrunning{L. Chen, et al.}

\institute{$^\dagger$ University of Rochester\space  \space  \space  \space \space  \space  \space  \space $^\ddag$ OPPO US Research Center  \email{lchen63@cs.rochester.edu} }

\maketitle
\begin{abstract}
When people deliver a speech, they naturally move heads, and this rhythmic head motion conveys prosodic information. However, generating a lip-synced video while moving head naturally is challenging.
While remarkably successful, existing works either generate still talking-face videos or rely on landmark/video frames as sparse/dense mapping guidance to generate head movements, which leads to unrealistic or uncontrollable video synthesis.
To overcome the limitations, we propose a 3D-aware generative network along with a hybrid embedding module and a non-linear composition module. Through modeling the head motion and facial expressions\footnote{In our setting, facial expression means facial movement(e.g., blinks, and lip \& chin movements).} explicitly, manipulating 3D animation carefully, and embedding reference images dynamically, our approach achieves controllable, photo-realistic, and temporally coherent talking-head videos with natural head movements. Thoughtful experiments on several standard benchmarks demonstrate that our method achieves significantly better results than the state-of-the-art methods in both quantitative and qualitative comparisons. The code is available on \url{https://github.com/lelechen63/Talking-head-Generation-with-Rhythmic-Head-Motion}.

\end{abstract}

\section{Introduction}
\label{sec:intro}



People naturally emit head movements when they speak, which contain non-verbal information that helps the audience comprehend the speech content~\cite{cassell1999speech,ginosar2019gestures}. Modeling the head motion and then generating a controllable talking-head video are valuable problems in computer vision. For example, it can benefit the research of adversarial attacks in security or provide more training samples for supervised learning approaches. Meanwhile, it is also crucial to real-world applications, such as enhancing speech comprehension for hearing-impaired people, and generating virtual characters with synchronized facial movements to speech audio in movies/games.


\begin{figure}[t]
\includegraphics[width= \linewidth]{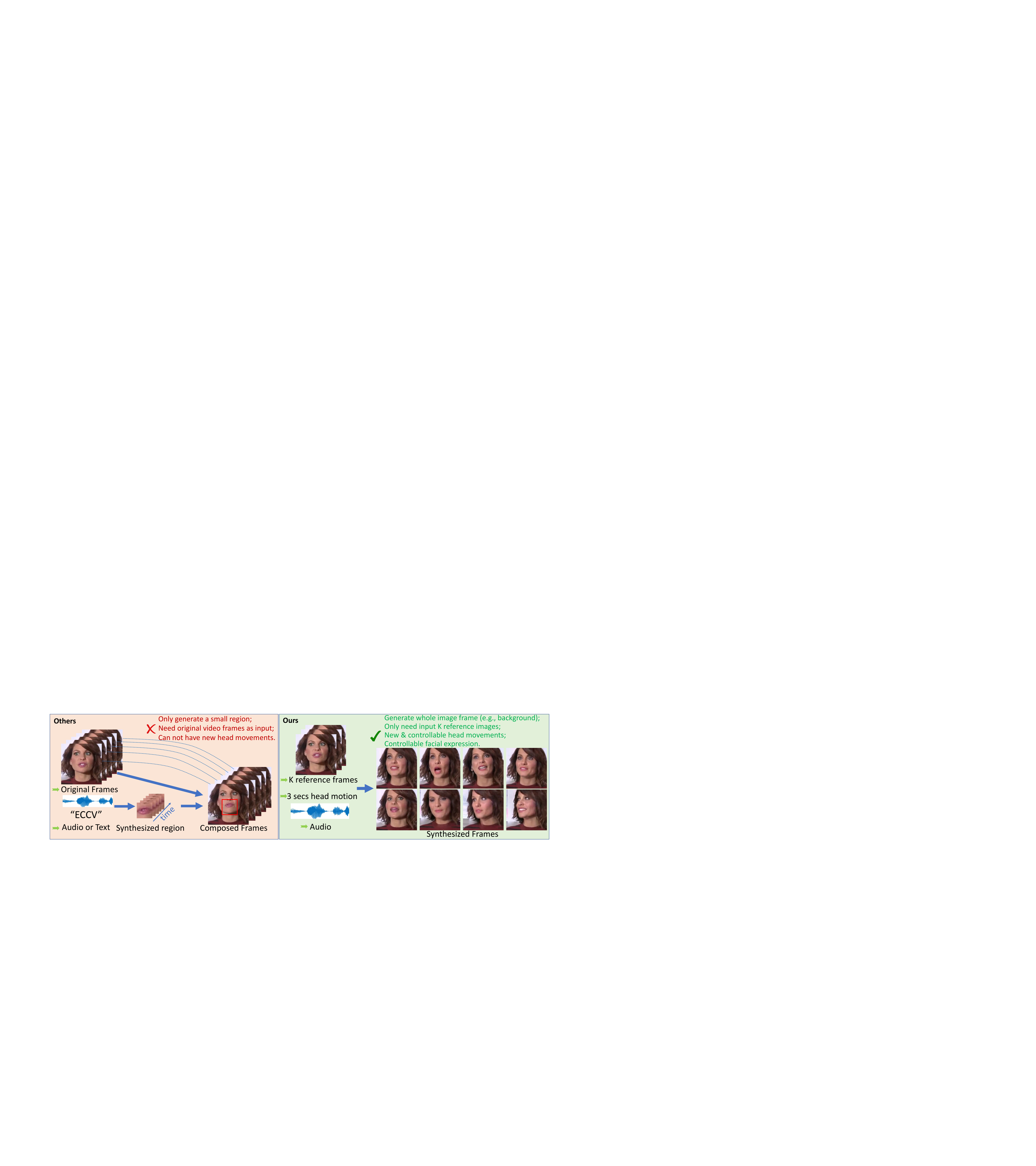}
\caption{The comparisons of other methods~\cite{suwajanakorn2017synthesizing,fried2019text} and our method. The green arrows denote inputs. ~\cite{suwajanakorn2017synthesizing,fried2019text} require whole original video frames as input and can only edit a small region (e.g., lip region) on the original video frames. In contrast, through learning the appearance embedding from $K$ reference frames and predicting future head movements using the head motion vector extracted from the 3-sec reference video, our method generates whole future video frames with controllable head movements and facial expressions.}
\label{fig:teaser}
\end{figure}

We humans can infer the visual prosody from a short conversation with speakers~\cite{munhall2004visual}. Inspired by that, in this paper, we consider such a task: given a short video \footnote{E.g., a 3-sec video. We use it to learn head motion and only sample $K$ images as reference frames.} of the target subject and an arbitrary reference audio, generating a photo-realistic and lip-synced talking-head video with natural head movements. Similar problems~\cite{chung2017you,zhou2019talking,ijcai2019-129,vougioukas2019realistic,chen2019hierarchical,wiles2018x2face,zakharov2019few,wang2018fewshotvid2vid} have been explored recently. However, several challenges on how to explicitly use the given video and then model the apparent head movements remain unsolved. In the rest of this section, we discuss our technical contributions concerning each challenge.


The deformation of a talking-head consists of his/her intrinsic subject traits, head movements, and facial expressions, which are highly convoluted. This complexity stems not only from modeling face regions but also from modeling head motion and background. While previous audio-driven talking-face generation methods~\cite{Chung18b,wiles2018x2face,zhou2019talking,chen2018lip,ijcai2019-129,chen2019hierarchical} could generate lip-synced videos, they omit the head motion modeling, thus can only generate still talking-face with expressions under a fixed facial alignment. Other landmark/image-driven generation methods~\cite{zakharov2019few,wang2018fewshotvid2vid,wang2018vid2vid,wang2018high,wiles2018x2face} can synthesize moving-head videos relying on the input facial landmarks or video frames as guidance to infer the convoluted head motion and facial expressions. However, those methods fail to output a controllable talking-head video (e.g., control facial expressions with speech), which greatly limits their use. To address the convoluted deformation problem, we propose a simple but effective method to decompose the head motion and facial expressions.     

Another challenge is exploiting the information contained in the reference image/video. While the few-shot generation methods~\cite{zakharov2019few,wang2018fewshotvid2vid,liu2019few,yoo2019coloring} can synthesize videos of unseen subjects by leveraging $K$ reference images, they only utilize global appearance information. There remains valuable information underexplored. For example, we can infer the individual's head movement characteristics by leveraging the given short video. Therefore, we propose a novel method to extrapolate rhythmic head motion based on the short input video and the conditioned audio, which enables generating a talking-head video with natural head poses. Besides, we propose a novel hybrid embedding network dynamically aggregating information from the reference images by approximating the relation between the target image with the reference images.

People are sensitive to any subtle artifacts and perceptual identity changes in a synthesized video, which are hard to avoid in GAN-based methods. 3D graphics modeling has been introduced by~\cite{fried2019text,kim2018deep,zhou2020rotate} in GAN-based methods due to its stability. In this paper, we employ the 3D modeling along with a novel non-linear composition module to alleviate the visual discontinuities caused by apparent head motion or facial expressions. 

Combining the above features, which are designed to overcome limitations of existing methods, our framework can generate a talking-head video with natural head poses that conveys the given audio signal. We conduct extensive experimental validation with comparisons to various state-of-the-art methods on several benchmark datasets (e.g., VoxCeleb2~\cite{Chung18b} and {LRS3-TED~\cite{Afouras18d}} datasets) under several different settings (e.g., audio-driven and landmark-driven). Experimental results show that the proposed framework effectively addresses the limitations of those existing methods.

\section{Related Work}
\label{sec:related}
\subsection{Talking-head Image Generation}
\label{subsec:talking-syn}
The success of graphics-based approaches has been mainly limited to synthesizing talking-head videos for a specific person~\cite{garrido2015vdub,bregler1997video,chang2005transferable,liu2011realistic,suwajanakorn2017synthesizing,fried2019text}. For instance, Suwajanakorn et al.~\cite{suwajanakorn2017synthesizing} generate a small region (e.g., lip region, see Fig.~\ref{fig:teaser})\footnote{ We intent to highlight the different between \cite{suwajanakorn2017synthesizing,fried2019text} and our work. While they generate high-quality videos, they can not disentangle the head motion and facial movement due to their intrinsic limitations.} and compose it with a retrieved frame from a large video corpus of the target person to produce photo-realistic videos. Although it can synthesize fairly accurate lip-synced videos, it requires a large amount of video footage of the target person to compose the final video. Moreover, this method can not be generalized to an unseen person due to the rigid matching scheme. More recently, video-to-video translation has been shown to be able to generate arbitrary faces from arbitrary input data. While the synthesized video conveys the input speech signal, the talking-face generation methods~\cite{pumarola2019ganimation,chung2017you,zhou2019talking,ijcai2019-129,vougioukas2019realistic,chen2019hierarchical} can only generate facial expressions without any head movements under a fixed alignment since the head motion modeling has been omitted. Talking-head methods~\cite{zakharov2019few,wang2018fewshotvid2vid,wiles2018x2face} can generate high-quality videos guided by landmarks/images. However, these methods can not generate controllable video since the facial expressions and head motion are convoluted in the guidance, e.g., using audio to drive the generation. In contrast, we explicitly model the head motion and facial expressions in a disentangled manner, and propose a method to extrapolate rhythmic head motion to enable generating future frames with natural head movements.  

\subsection{Related Techniques}
\label{subsec:related_tec}
\noindent \textbf{Embedding Network} refers to the external memory module to learn common feature over few reference images of the target subjects. And this network is critical to identity-preserving performance in generation task. Previous works~\cite{chung2017you,chen2019hierarchical,vougioukas2019realistic,ijcai2019-129} directly use CNN encoder to extract the feature from reference images, then concatenate with other driven vectors and decode it to new images. However, this encoder-decoder structure {suffers identity-preserving problem caused by the deep convolutional layers}. Wiles et al.~\cite{wiles2018x2face} propose an embedding network to learn a bilinear sampler to map the reference frames to a face representation. More recently, \cite{zakharov2019few,yoo2019coloring,wang2018fewshotvid2vid} compute part of the network weights dynamically based on the reference images, which can adapt to novel cases quickly. Inspired by~\cite{wang2018fewshotvid2vid}, we propose a hybrid embedding network to aggregate appearance information from reference images with apparent head movements. 

\noindent \textbf{ Image matting function} has been well explored in image/video generation task~\cite{vondrick2016generating,wang2018vid2vid,pumarola2019ganimation,chen2019hierarchical,wang2018fewshotvid2vid}. For instance, Pumarola et al.~\cite{pumarola2019ganimation} compute the final output image by $\hat{\mathbf{I}} = (1 - \mathbf{A}) * \mathbf{C} + \mathbf{A} * \mathbf{I}_r $, where $\mathbf{I}_r$, $\mathbf{A}$ and $\mathbf{C}$ are input reference image, attention map and color mask, respectively. The attention map $\mathbf{A}$ indicates to what extend each pixel of $\mathbf{I}_r$ contributes to the output image $\hat{\mathbf{I}}$. However, this attention mechanism may not perform well if there exits a large deformation between reference frame $\mathbf{I}_r$ and target frame $\hat{\mathbf{I}}$. Wang et al.~\cite{wang2018high} use estimated optic flow to warp the $\mathbf{I}_r$ to align with $\hat{\mathbf{I}}$, which is computationally expensive and can not estimate the rotations in talking-head videos. In this paper, we propose a 3D-aware solution along with a non-linear composition module to better tackle the misalignment problem caused by apparent head movements. 

\section{Method}
\label{sec:method}
\subsection{Problem Formulation}
\label{subsec:problem_formulation}
We introduce a neural approach for talking-head video generation, which takes as input sampled video frames, ${\mathbf{y}}_{1:\tau} \equiv {\mathbf{y}}_1,...,{\mathbf{y}}_{\tau}$, of the target subject and a sequence of driving audio, ${\mathbf{x}}_{\tau+ 1:T}  \equiv {\mathbf{x}}_{\tau + 1},...,{\mathbf{x}}_T$, and synthesizes target video frames, $\hat{\mathbf{y}}_{\tau + 1:T}  \equiv \hat{\mathbf{y}}_{\tau + 1}, ...,\hat{\mathbf{y}}_T$, that convey the given audio signal with realistic head movements. To explicitly model facial expressions and head movements, we decouple the full model into three sub-models: a facial expression learner ($\Psi$), a head motion learner ($\Phi$), and a 3D-aware generative network ($\Theta$). Specifically, given an audio sequence ${\mathbf{x}}_{\tau+1:T}$ and an example image frame ${\mathbf{y}}_{t_{\mathcal{M}}}$, $\Psi$ generates the facial expressions $\hat{\mathbf{p}}_{\tau+1: T}$. Meanwhile, given a short reference video ${\mathbf{y}}_{1:\tau}$ and the driving audio ${\mathbf{x}}_{\tau+1:T}$, $\Phi$ extrapolates natural head motion $\hat{\mathbf{h}}_{\tau + 1:T}$ to manipulate the head movements. Then, the $\Theta$ generates video frames $\hat{\mathbf{y}}_{\tau + 1:T}$ using $\hat{\mathbf{p}}_{\tau+1: T}$, $\hat{\mathbf{h}}_{\tau+1: T}$, and ${\mathbf{y}}_{1:\tau}$. Thus, the full model is given by:
\begin{align*}
\hat{\mathbf{y}}_{ \tau+1:T} 
= \Theta(\mathbf{y}_{1:\tau}, \hat{\mathbf{p}}_{\tau+1: T}, \hat{\mathbf{h}}_{\tau+1: T})
=  \Theta( \mathbf{y}_{1:\tau},\Psi( \mathbf{y}_{t_{\mathcal{M}}} , \mathbf{x}_{\tau +1:T} ), \Phi(\mathbf{y}_{1:\tau},\mathbf{x}_{\tau +1:T} )  )  \enspace.
\end{align*}

\begin{figure}[t]
\includegraphics[width=1.0 \linewidth]{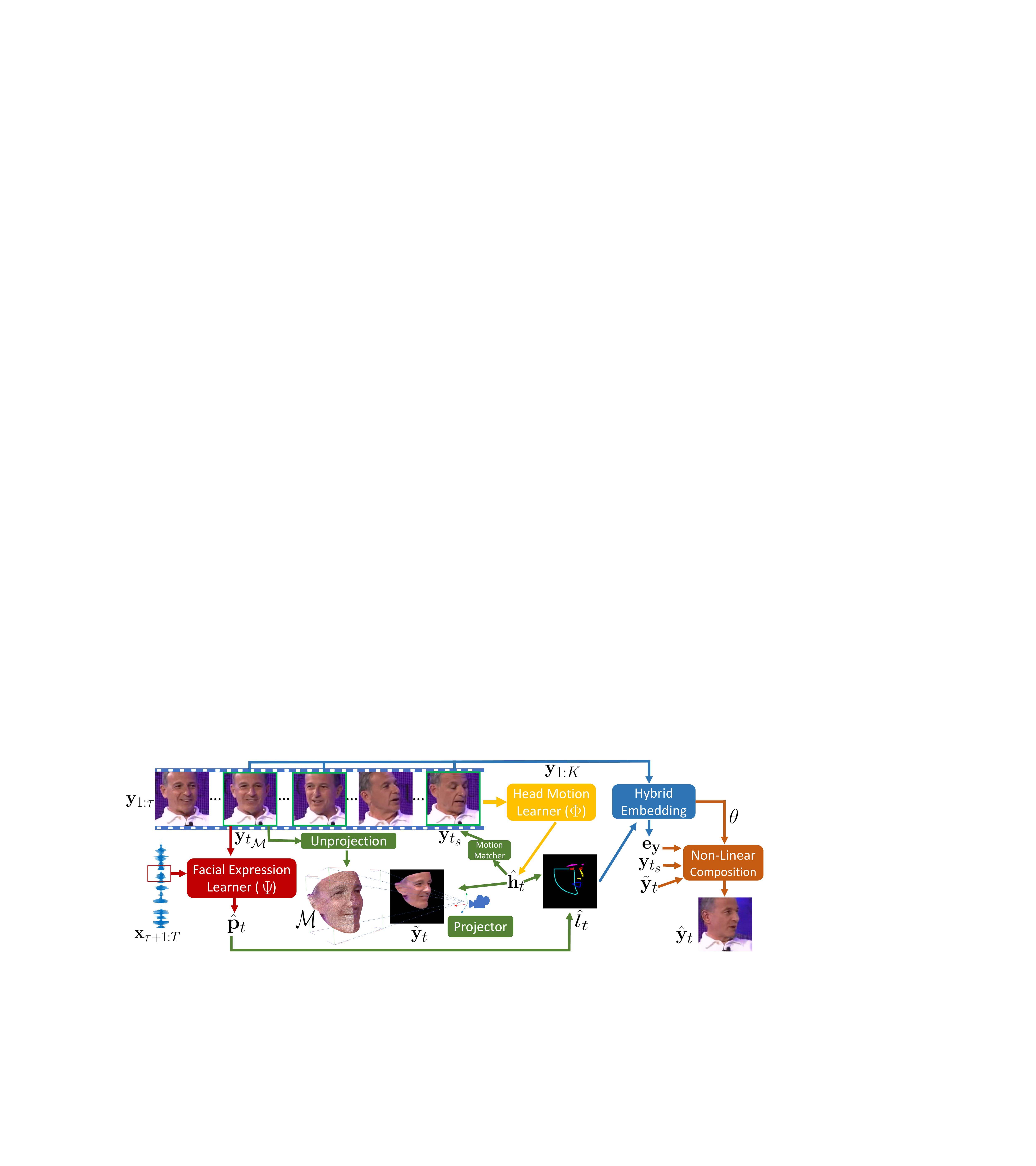}
\caption{The overview of the framework. $\Psi$ and $\Phi$ are introduced in Sec.~\ref{subsec:facial_exp} and Sec.~\ref{subsec:head_mo}, respectively. The 3D-aware generative network consists of a 3D-Aware module (\textcolor{OliveGreen}{green} part, see Sec.~\ref{subsec:3d_man}), a hybrid embedding module (\textcolor{RoyalBlue}{blue} part, see Sec.~\ref{subsec:hybrid}), and a non-linear composition module (\textcolor{Orange}{orange} part, see Sec~\ref{subsec:non-linear}).}
\label{fig:main}
\end{figure}

The proposed framework (see Fig.~\ref{fig:main}) aims to exploit the facial texture and head motion information in $\mathbf{y}_{1:\tau}$ and maps the driving audio signal to a sequence of generated video frames $\hat{\mathbf{y}}_{ \tau+1:T}$. 

\subsection{The Facial Expression Learner}
\label{subsec:facial_exp}
The facial expression learner $\Psi$ (see Fig.~\ref{fig:main} red block) receives as input the raw audio $\mathbf{x}_{\tau +1:T}$, and a subject-specific image $\mathbf{y}_{t_{\mathcal{M}}}$ (see Sec.~\ref{subsec:3d_man} about how to select $\mathbf{y}_{t_{\mathcal{M}}}$), from which we extract the landmark identity feature $\mathbf{p}_{t_{\mathcal{M}}}$. The desired outputs are PCA components ${\hat{\mathbf{p}}}_{\tau + 1 :T}$ that convey the facial expression. During training, we mix $\mathbf{x}_{\tau +1:T}$ with a randomly selected noise file in 6 to 30 dB SNRs with 3 dB increments to increase the network robustness. At time step $t$, we encode audio clip $\mathbf{x}_{t-3:t+4}$ ($0.04 \times 7$ Sec) into an audio feature vector, concatenate it with the encoded reference landmark feature, and then decode the fused feature into PCA components of target expression ${\hat{\mathbf{p}}}_{t}$. During inference, we add the identity feature $\mathbf{p}_{t_{\mathcal{M}}}$ back to the resultant facial expressions ${\hat{\mathbf{p}}}_{\tau +1 : T}$ to keep the identity information. 
\begin{figure}
\includegraphics[width=0.98 \linewidth]{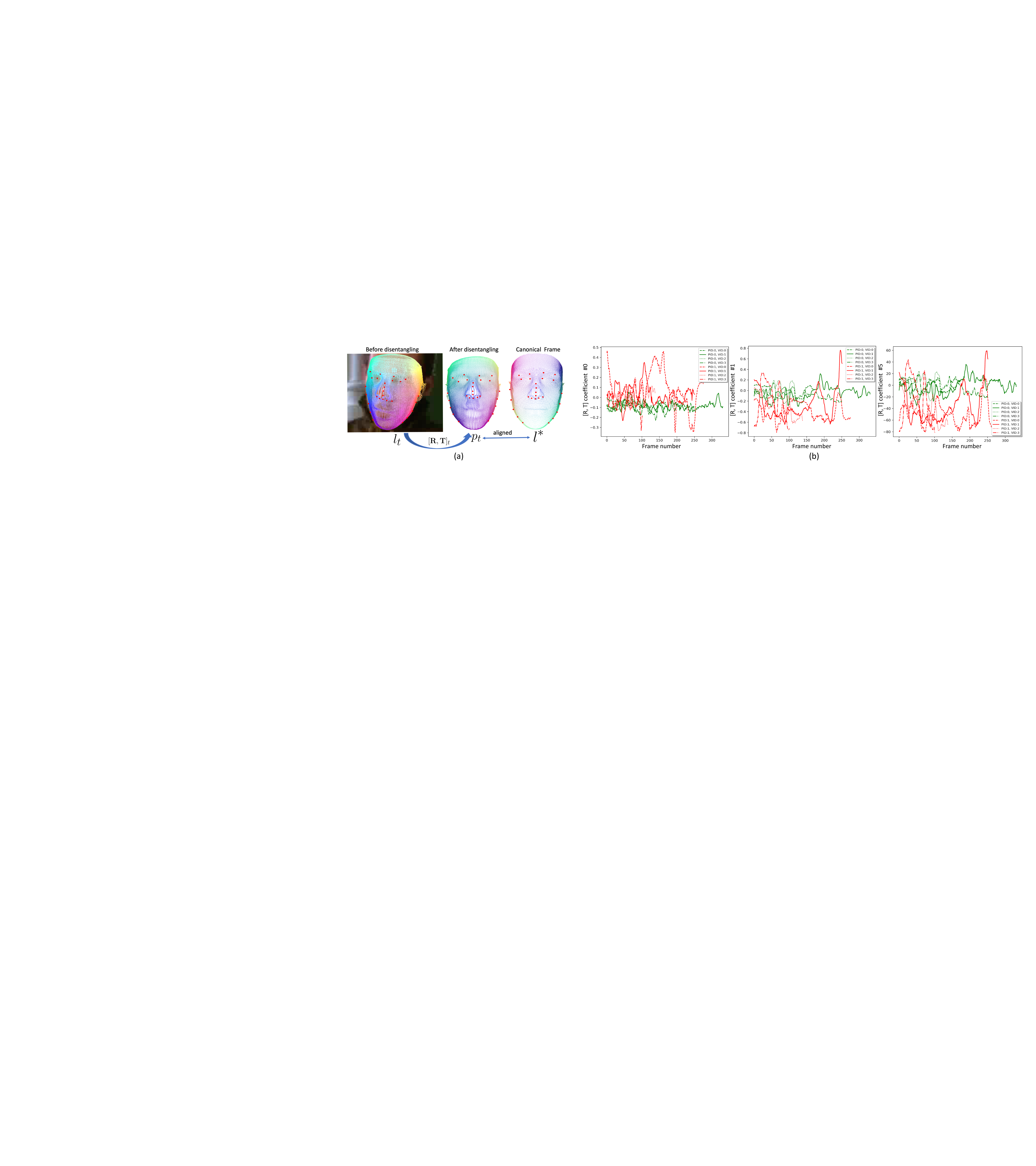}
\caption{ (a) shows the motion disentangling process. To avoid the noise caused by non-rigid deformation, we select 27 points (red points), which contain less non-rigid deformation. (b) shows the head movement coefficients of two randomly-selected identities (each with four different videos). PID, VID indicate identity id and video id, respectively.}
\label{fig:rt_compute}
\end{figure}

\subsection{The Head Motion Learner}
\label{subsec:head_mo}
To generate talking-head videos with natural head movements, we introduce a head motion learner $\Phi$, which disentangles the reference head motion ${\mathbf{h}}_{1:\tau}$ from the input reference video clip $\mathbf{y}_{1:\tau}$ and then predicts the head movements $\hat{\mathbf{h}}_{\tau +1:T}$ based on the driving audio $\mathbf{x}_{1:T}$ and the disentangled motion ${\mathbf{h}}_{1:\tau}$.

\noindent {\bf{Motion Disentangler.}}\indent Rather than disentangling the head motion problem in image space, we learn the head movements $\mathbf{h}$ in {3D geometry space} (see Fig.~\ref{fig:rt_compute}(a)). Specifically, we compute the transformation matrix $[\mathbf{R},\mathbf{T}] \in \mathbb{R}^{6}$ to represent $\mathbf{h}$, where we omit the camera motion and other factors. At each time step $t$, we compute the rigid transformation $[\mathbf{R},\mathbf{T}]_t$ between $\mathbi{l}_t$ and canonical 3D facial landmark $\mathbi{l}^*$, which is ${\mathbi{l}^*} \approx \mathbf{p}_t =  \mathbf{R}_t  \mathbi{l}_t + \mathbf{T}_t$, where $\approx$ denotes aligned. After transformation, the head movement information is removed, and the resultant ${\mathbf{p}}_t$ only carries the facial expressions information.
 
\noindent {\bf{The natural head motion extrapolation.}}\indent We randomly select two subjects (each with four videos) from VoxCeleb2~\cite{Chung18b} and plot the motion coefficients in Fig.~\ref{fig:rt_compute}(b). We can find that one subject (same PID, different VID) has similar head motion patterns in different videos, and different subjects (different PID) have much different head motion patterns. To explicitly learn the individual's motion from ${\mathbf{h}}_{1:\tau}$ and the audio-correlated head motion from ${{\mathbf{x}}}_{1:T}$, we propose a few-shot temporal convolution network $\Phi$, which can be formulated as: ${\hat{\mathbf{h}}}_{\tau+1:T} = \Phi( f({{\mathbf{h}}}_{1:\tau}, {{\mathbf{x}}}_{1:\tau}), {{\mathbf{x}}}_{\tau+1:T})$. In details, the encoder $f$ first encodes ${\mathbf{h}}_{1:\tau}$ and ${{\mathbf{x}}}_{1:\tau}$ to a high-dimensional reference feature vector, and then transforms the reference feature to network weights $\textbf{w}$ using a multi-layer perception. Meanwhile, ${{\mathbf{x}}}_{\tau+1:T}$ is encoded by several temporal convolutional blocks. Then, the encoded audio feature is processed by a multi-layer perception, where the weights are dynamically learnable $\textbf{w}$, to generate ${\hat{\mathbf{h}}}_{\tau+1:T}$.
 
\subsection{The 3D-Aware Generative Network}
\label{subsec:3dgenerator}
To generate controllable videos, we propose a 3D-aware generative network $\Theta$, which fuses the head motion ${\mathbf{h}}_{\tau + 1:T}$, and facial expressions ${\mathbf{p}}_{\tau + 1:T}$ with the appearance information in ${\mathbf{y}}_{1:\tau }$ to synthesise target video frames, $\hat{\mathbf{y}}_{\tau + 1:T}$. The $\Theta$ consists of three modules: a \textit{3D-Aware} module, which manipulates the head motion to reconstruct an intermediate image that carries desired head pose using a differentiable unprojection-projection step to bridge the gap between images with different head poses. Comparing to landmark-driven methods~\cite{zakharov2019few,wang2018fewshotvid2vid,wang2018high}, it can alleviate the overfitting problem and make the training converge faster; a \textit{Hybrid Embedding} module, which is proposed to explicitly embed texture features from different appearance patterns carried by different reference images; a \textit{Non-Linear Composition} module, which is proposed to stabilize the background and better preserve the individual's appearance. Comparing to image matting function~\cite{pumarola2019ganimation,zhou2019talking,zakharov2019few,wang2018high,chen2019hierarchical}, our non-linear composition module can synthesize videos with natural facial expression, smoother transition, and a more stable background.

\section{3D-Aware Generation}
\label{sec:3dgan}
\subsection{\textit{\textbf{3D-Aware}} Module}
\label{subsec:3d_man}

{ \noindent {\bf{3D Unprojection}}\indent We assume that the image with frontal face contains more appearance information. Given a short reference video clip $\mathbf{y}_{1:\tau}$, we compute the head rotation $\mathbf{R}_{1:\tau}$ and choose the most visible frame $\mathbf{y}_{t_{\mathcal{M}}}$ with minimum rotation, such that $\mathbf{R}_{t_{\mathcal{M}}} \rightarrow 0$. Then we feed it to an unprojection network to obtain a 3D mesh $\mathcal{M}$. The Unprojection network is a U-net structure network~\cite{feng2018joint} and we pretrain it on 300W-LP~\cite{zhu2016face} dataset. After pretrianing, we fix the weights and use it to transfer the input image $\mathbf{y}_{t_{\mathcal{M}}}$ into the textured 3D mesh $\mathcal{M}$. The topology of $\mathcal{M}$ will be fixed for all frames in one video, we denote it as $\mathcal{F}^{\mathcal{M}}$.}
 
\noindent {\bf{3D Projector}}\indent  In order to get the projected image $\tilde{\mathbf{y}}_t$ from the 3D face mesh $\mathcal{M}$, we need to compute the correct pose for $\mathcal{M}$ at time $t$. Suppose $\mathcal{M}$ is reconstructed from frame $\mathbf{y}_{t_{\mathcal{M}}}, (1\le t_{\mathcal{M}}\le \tau)$, the position of vertices of $\mathcal{M}$ at time $t$ can be computed by:
\begin{equation}
\begin{aligned}
\mathcal{V}^{\mathcal{M}}_t = \mathbf{R}_t^{-1}(\mathbf{R}_{t_{\mathcal{M}}}\mathcal{V}^{\mathcal{M}}_{t_{\mathcal{M}}} + \mathbf{T}_{t_{\mathcal{M}}} - \mathbf{T}_t) \enspace,
\end{aligned}
\label{eq:projector}    
\end{equation}
where $\mathcal{V}^{\mathcal{M}}_{t_{\mathcal{M}}}$ denotes the position of vertices of $\mathcal{M}$ at time $t_{\mathcal{M}}$. Hence, $\mathcal{M}$ at time $t$ can be represented as $\left[\mathcal{V}^{\mathcal{M}}_t, \mathcal{F}^{\mathcal{M}}\right]$. Then, a differentiable rasterizer~\cite{liu2019softras} is applied here to render $\tilde{\mathbf{y}}_t$ from $\left[\mathcal{V}^{\mathcal{M}}_t, \mathcal{F}^{\mathcal{M}}\right]$. After rendering, $\tilde{\mathbf{y}}_t$ carries the same head pose as the target frame ${\mathbf{y}}_{t}$ and the same expression as ${\mathbf{y}}_{t_{\mathcal{M}}}$. 

\noindent {\bf{Motion Matcher}}\indent We randomly select $K$ frames $\mathbf{y}_{1:K}$ from $\mathbf{y}_{1:\tau}$ as reference images. To infer the prior knowledge about the background of the target image from $\mathbf{y}_{1:K}$, we propose a motion matcher, $M({\mathbf{h}}_t, \mathbf{h}_{1:K}, \mathbf{y}_{1:K})$, which matches a frame $\mathbf{y}_{t_s}$ from $\mathbf{y}_{1:K}$ with the nearest background as $\mathbf{y}_{t}$ by comparing the similarities between $\mathbf{h}_{t}$ with $\mathbf{h}_{1:K}$. To solve this background retrieving problem, rather than directly computing the similarities between facial landmarks, we compute the geometry similarity cost $c_{t}$ using the head movement coefficients and choose the $\mathbf{y}_{t_s}$ with the least cost $c_t$. That is:
\begin{equation}
\begin{aligned}
c_{t}= \min_{k=1,2,...K}  \lVert \mathbf{h}_t - \mathbf{h}_k \rVert^2_2 \enspace.
\end{aligned}
\label{eq:cost}    
\end{equation}
The matched $\mathbf{y}_{t_s}$ is passed to the non-linear composition module (see Sec.~\ref{subsec:non-linear}) since it carries the similar background pattern as ${\mathbf{y}}_t$. During training, we manipulate the selection of $\mathbf{y}_{t_s}$ by increasing the cost $c_{t}$, which makes the non-linear composition module more robust to different $\mathbf{y}_{t_s}$.  
 
 \begin{figure}[t]
\includegraphics[width= \linewidth]{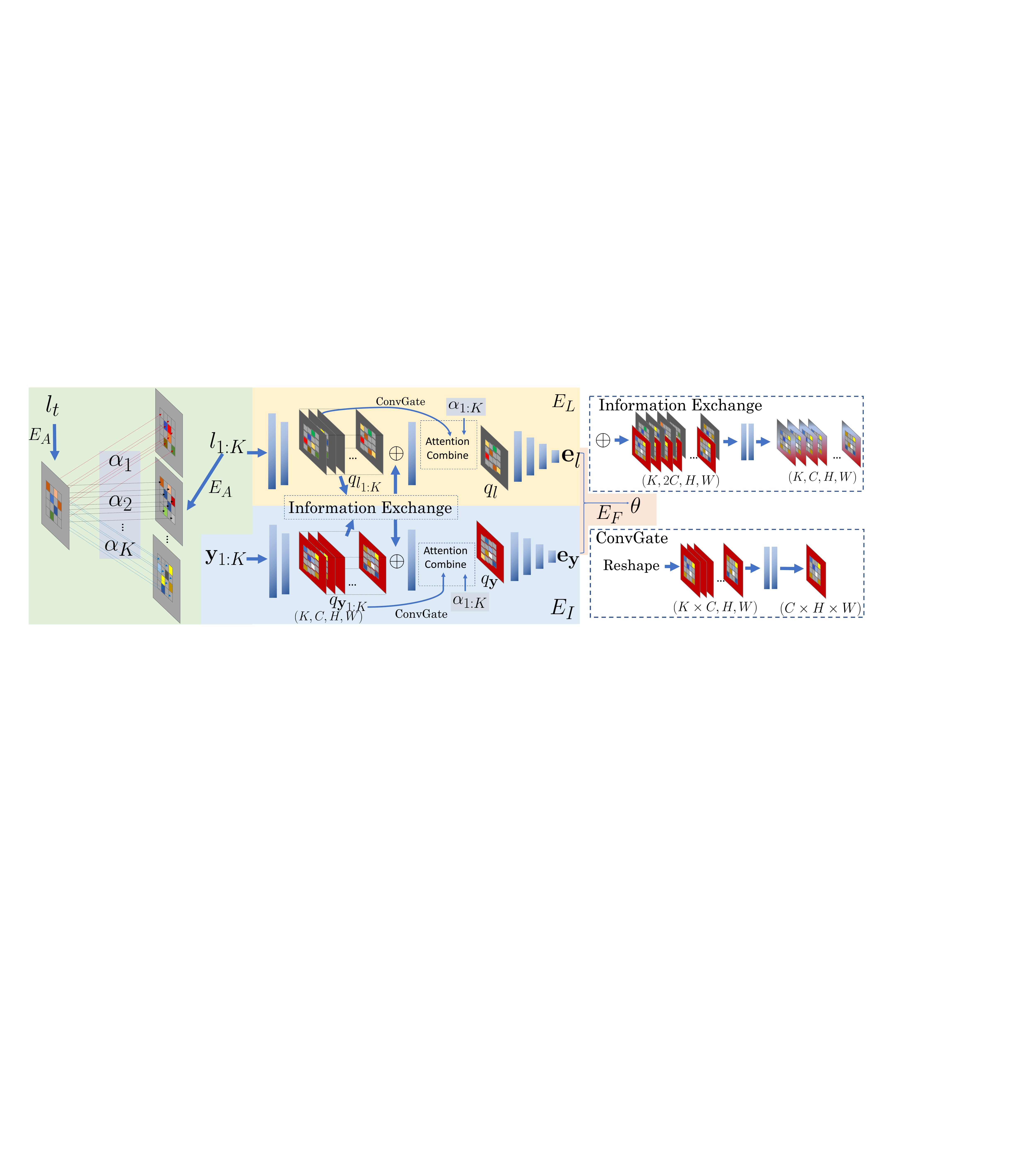}
\caption{The network structure of hybrid embedding module. Green, yellow, blue, pink parts are activation encoder $E_A$, landmark encoder $E_L$, image encoder $E_I$, and fusion network $E_F$, respectively. Inside information exchange block, we first concatenate features from two modality, and then apply two bottleneck layer to decrease the channel dimension. Inside ConvGate block, we first reshape the feature shape and then apply two bottleneck layer to decrease the channel dimension. The attention combine is shown in Eq.~\ref{eq:qy_combine}. }
\label{fig:attention}
\end{figure}

\subsection{Hybrid Embedding Module}
\label{subsec:hybrid}
Instead of averaging the $K$ reference image features~\cite{zakharov2019few,chung2017you,wiles2018x2face}, we design a hybrid embedding mechanism to dynamically aggregate the $K$ visual features.

Recall that we obtain the facial expression ${\mathbf{p}}_{t}$ (Sec.~\ref{subsec:facial_exp}) and the head motion ${\mathbf{h}}_{t}$ (Sec.~\ref{subsec:head_mo}) for time $t$. Here, we combine them to landmark ${\mathbi{l}}_{t}$ and transfer it to a gray-scale image. Our hybrid embedding network (Fig.~\ref{fig:attention}) consists of four sub-networks: an Activation Encoder ($E_{A}$, green part) to generate the activation map between the query landmark $\mathbi{l}_t$ and reference landmarks $\mathbi{l}_{1:K}$; an Image Encoder ($E_{I}$, blue part) to extract the image feature $\mathbf{e}_{\mathbf{y}}$; a Landmark Encoder ($E_{L}$, yellow part) to extract landmark feature $\mathbf{e}_{\mathbi{l}}$; a fusion network ($E_{F}$, pink part) to aggregate image feature and landmark feature into parameter set $\theta$. The overview of the embedding network is performed by:
\begin{align}
\label{eq:attention}
&\boldsymbol{\alpha}_{k} = \text{softmax}( E_{A}(\mathbi{l}_k) \odot E_{A}(\mathbi{l}_t)) \enspace, \enspace  \enspace k \in [1,K]
\enspace, \\
\label{eq:img_embed}
&\mathbf{e}_{\mathbf{y}} =  E_{I}(\mathbf{y}_{1:K}, \boldsymbol{\alpha}_{1:K})
\enspace, \enspace \enspace  \mathbf{e}_{\mathbi{l}} =  E_{L}(\mathbi{l}_{1:K}, \boldsymbol{\alpha}_{1:K})   \enspace, \\
\label{eq:fusion}
&\theta  = E_F(\mathbf{e}_{\mathbi{l}}, \mathbf{e}_{\mathbf{y}}) \enspace, 
\end{align}
where $\odot$ denotes element-wise multiplication. We regard the activation map $\boldsymbol{\alpha}_{k}$ as the activation energy between ${\mathbi{l}_k}$ and ${\mathbi{l}_t}$, which approximates the similarity between ${\mathbf{y}_k}$ and ${\mathbf{y}_t}$. 

We observe that different referent images may carry different appearance patterns, and those appearance patterns share some common features with the reference landmarks (e.g., head pose, and edges). Assuming that knowing the information of one modality can better encode the feature of another modality~\cite{tian2018audio}, we hybrid the information between $E_{I}$ and $E_{L}$. Specifically, we use two convolutional layers to extract the feature map $\mathbf{q}_{\mathbf{y}_{1:K}}$ and $\mathbf{q}_{\mathbi{l}_{1:K}}$, and then forward them to an information exchange block to refine the features. After exchanging the information, the hybrid feature is concatenated with the original feature $\mathbf{q}_{\mathbf{y}_{1:K}}$. Then, we pass the concatenated feature to a bottleneck convolution layer to produce the refined feature $\mathbf{q}'_{\mathbf{y}_{1:K}}$. Meanwhile, we apply a ConvGate block on $\mathbf{q}_{\mathbf{y}_{1:K}}$ to self-aggregate the features from $K$ references, and then combine the gated feature with the refined feature using learnable activation map $\boldsymbol{\alpha}_{k}$. This attention combine step can be formulated as:
\begin{align}
\label{eq:qy_combine}
\mathbf{q}_{\mathbf{y}}= \sum_{i=1}^{K} (\boldsymbol{\alpha}_{1:K} \odot \mathbf{q}'_{\mathbf{y}_{1:K}}) + \text{ConvGate}(\mathbf{q}_{\mathbf{y}_{1:K}})
\enspace.
\end{align}
Then, by applying several convolutional layers to the aggregated feature $\mathbf{q}_{\mathbf{y}}$, we obtain the image feature vector $\mathbf{e}_{\mathbf{y}}$. Similarly, the landmark feature vector $\mathbf{e}_{\mathbi{l}}$ can also be produced.

The Fusion network $E_{F}$ consists of three blocks: a $N$-layer image feature encoding block $E_{F_{I}}$, a landmark feature encoding block $E_{F_L}$, and a multi-layer perception block $E_{F_{P}}$. Thus, we can rewrite Eq.~\ref{eq:fusion} as:
\begin{align}
\label{eq:fusion2}
\{\theta_\gamma^i, \theta_\beta^i, \theta_S^i\}_{i \in [1,N]} = E_{F_P}({\text{softmax}}(  E_{F_L}(\mathbf{e}_{\mathbi{l}})) \odot E_{F_I}(\mathbf{e}_{\mathbf{y}}) )
\enspace,
\end{align}
where $\{\theta_\gamma^i, \theta_\beta^i, \theta_S^i\}_{i \in [1,N]}$ are the learnable network weights in the non-linear composition module in Sec.~\ref{subsec:non-linear}.

\begin{figure}
\includegraphics[width= \linewidth]{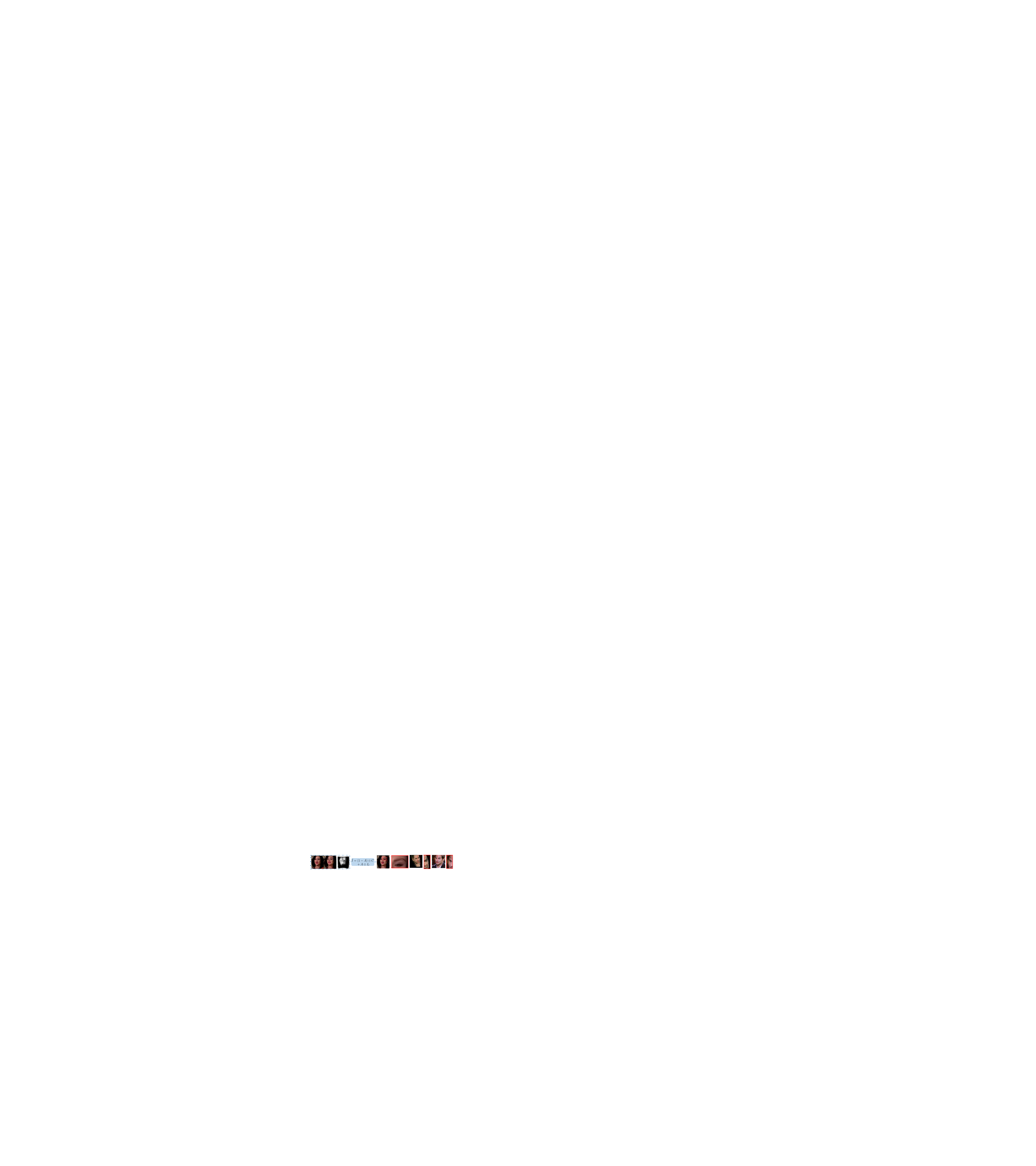}
\centering
\caption{The artifact examples produced by the image matting function. The blue dashed box shows the color map $\mathbf{C}$, reference image $\mathbf{I_r}$, and attention map $\mathbf{A}$ , respectively. The blue box shows the image matting function (see Sec.~\ref{subsec:related_tec}) followed by the final result with its zoom-in artifact area. Two other artifact examples are on the right side.}
\label{fig:artifact}
\end{figure}

\subsection{Non-Linear Composition Module} 
\label{subsec:non-linear}
\begin{figure}[t]
\includegraphics[width= \linewidth]{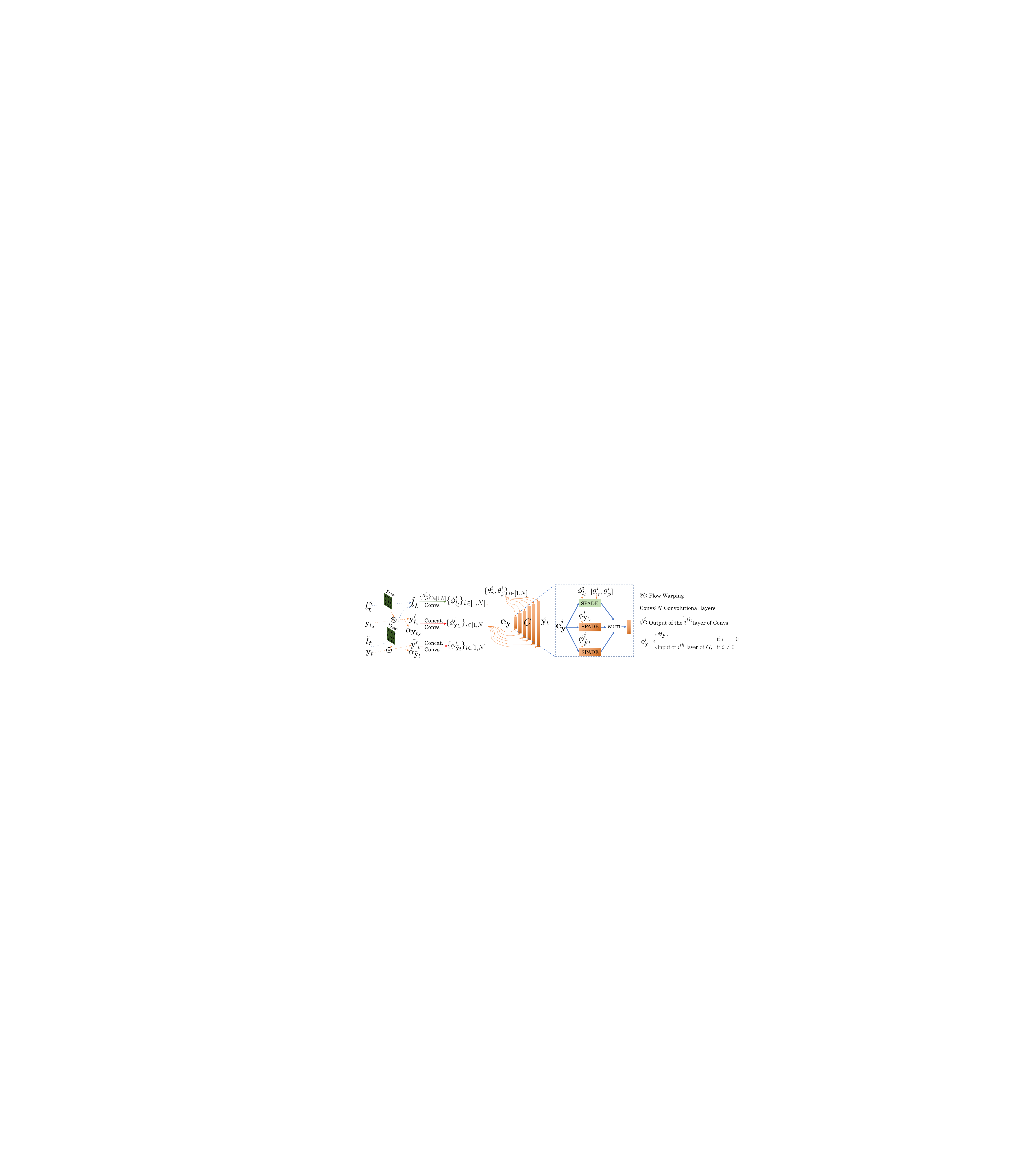}
\centering
\caption{The network structure of the non-linear composition module. The network weights $\{\theta^i_{S}, \theta^i_{\gamma}, \theta^i_{\beta}\}_{i \in [1,N]}$ and $\mathbf{e_y}$ are learned in embedding network (Sec.~\ref{subsec:hybrid}). To better encode ${\hat{\mathbi{l}}}_t$ with extra knowledge learned from the embedding network, we leverage $\{\theta^i_S\}_{i\in [1,N]}$ as the weights of the convolutional layers (green arrow) and save all the intermediate features $\{\phi^i_{\mathbi{l}_t}\}_{i \in [1,N]}$. Similarly, we obtain $\{\phi^i_{\tilde{\mathbf{y}}_t}, \phi^i_{{\mathbf{y}}_{t_s}}\}_{i \in [1,N]}$ with fixed convolutional weights (red arrow). Then, we input $\{\phi^i_{\tilde{\mathbf{y}}_t}, \phi^i_{{\mathbf{y}}_{t_s}}, \phi^i_{\mathbi{l}_t} \}_{i \in [1,N]}$ into the corresponding block of the decoder $G$ to guide the image generation.}
\label{fig:composite}
\end{figure}

Since the projected image $\tilde{\mathbf{y}}_t$ carries valuable facial texture with better alignment to the ground-truth, and the matched image $\mathbf{y}_{t_s}$ contains similar background pattern, we input them to our composition module to ease the burden of the generation. We use FlowNet~\cite{wang2018vid2vid} to estimate the flow between ${\tilde{\mathbi{l}}}_t$ and ${\hat{\mathbi{l}}}_t$, and warp the projected image to obtain $\tilde{\mathbf{y}}_t'$. Meanwhile, the FlowNet also outputs an attention map $\boldsymbol{\alpha}_{\tilde{y}_t}$. Similarly, we obtain the warped matched image and its attention map $[\mathbf{y'}_{t_s}, \boldsymbol{\alpha}_{{\mathbf{y}}_{t_s}}]$. Although image matting function is a popular way to combine the warped image and the raw output of the generator~\cite{wang2018fewshotvid2vid,wang2018vid2vid,chen2019hierarchical,pumarola2019ganimation} using these attention maps, it may generate apparent artifacts (Fig.~\ref{fig:artifact}) caused by misalignment, especially in the videos with apparent head motion. To solve this problem, we introduce a nonlinear combination module (Fig.~\ref{fig:composite}).


The decoder $G$ consists of N layers and each layer contains three parallel SPADE blocks~\cite{park2019semantic}. Inspired by~\cite{wang2018fewshotvid2vid}, in the $i^{th}$ layer of G, we first convolve $[\theta_\gamma^i,\theta_\beta^i]$ with $\phi_{l_t}^i$ to generate the scale and bias map $\{\gamma^i, \beta^i\}_{\phi_{l_t}^i}$ of SPADE (green box) to denormalize the appearance vector $\mathbf{e}_{\mathbf{y}}^i$. This step aims to sample the face representation $\mathbf{e}_{\mathbf{y}}^i$ towards the target pose and expression. In other two SPADE blocks (orange box), the scale and bias map $\{\gamma^i,\beta^i\}_{\phi_{{{\mathbf{y}}}_{t_s}}, \phi_{\tilde{\mathbf{y}}_t}}$ are generated by fixed weights convoluted on $\phi_{{\mathbf{y}}_{t_s}}$ and $\phi_{\tilde{{\mathbf{y}}}_t}$, respectively. Then, we sum up the three denomalized features and upsample them with a transposed convolution. We repeat this non-linear combination module in all layers of G and apply a  hyperbolic tangent activation layer at the end of G to output fake image $\hat{\mathbf{y}}_t$.
Our experiment results (ablation study in Sec.~\ref{sec:quanti_res}) show that this parallel non-linear combination is more robust to the spatial-misalignment among the input images, especially when there is an apparent head motion.

\subsection{Objective Function}
\label{subsec:loss}

The Multi-Scale Discriminators~\cite{wang2018high} are employed to differentiate the real and synthesized video frames. The discriminators $D_1$ and $D_2$ are trained on two different input scales, respectively. Thus, the best generator $G^*$ is found by solving the following optimization problem:
\begin{equation}
\begin{aligned}
& G^* = \argminA_G\Big{(} \max_{D_1,D_2}\sum_{k=1,2} \mathcal{L}_{\text{GAN}}(G,D_k) \\
& + \lambda_{\text{FM}} \sum_{k=1,2} \mathcal{L}_{\text{FM}}(G,D_k) \Big{)} + \lambda_{\text{PCT}} \mathcal{L}_{\text{PCT}}(G) + \lambda_{W}\mathcal{L}_{W}(G)  \enspace,
\end{aligned}
\label{eq:loss}    
\end{equation}

where $\mathcal{L}_{\text{FM}}$ and $\mathcal{L}_{W}$ are the feature matching loss and flow loss proposed in~\cite{wang2018vid2vid}. The $\mathcal{L}_{\text{PCT}}$ is the VGG19~\cite{Simonyan15} perceptual loss term, which measures the perceptual similarity. The $\lambda_{\text{FM}}$, $\lambda_{\text{PCT}}$, and $\lambda_{W}$ control the importance of the four loss terms. 

\section{Experiments Setup}
\label{sec:experiments}

\noindent \textbf{Dataset.}\quad We quantitatively and qualitatively evaluate our approach on three datasets: LRW~\cite{Chung16}, VoxCeleb2~\cite{Chung18b}, and LRS3-TED~\cite{Afouras18d}. The LRW dataset consists of 500 different words spoken by hundreds of different speakers in the wild and the VoxCeleb2 contains over 1 million utterances for 6,112 celebrities. The videos in LRS3-TED~\cite{Afouras18d} contain more diverse and challenging head movements than the others. We follow same data split as \cite{Chung18b,Chung16,Afouras18d}.

\noindent \textbf{Implementation Details}\footnote{All experiments are conducted on an NVIDIA DGX-1 machine with 8 32GB V100 GPUs. It takes 5 days to converge the training on VoxCeleb2/LRS3-TED since they contain millions of video clips. It takes less than 1 day to converge the training on the GRID/CREMA. For more details, please refer to supplemental materials.}. \quad We follow the same training procedure as \cite{wang2018vid2vid}. We adopt ADAM~\cite{KingmaB14} optimizer with $\text{lr} = 2 \times 10^{-4}$ and $(\beta_1,\beta_2) = (0.5,0.999)$. During training, we select $K = 1,8,32$ in our embedding network. The $\tau$ in all experiments is 64. The $\lambda_{\text{FM}} = \lambda_{\text{PCT}} = \lambda_{\text{PCT}} = 10$ in Eq.~\ref{eq:loss}. 

\begin{table}[t]
  \centering
 \caption{Quantitative results of different audio to video methods on LRW dataset and VoxCeleb2 dataset. Our model mentioned in this table are trained from scratch. We bold each leading score.}
\begin{tabular*}{0.92\linewidth}{ l| c c c c |c c c c}
     \toprule
      \hline
Method  & \multicolumn{8}{c}{Audio-driven}  \\ \hline
Dataset & \multicolumn{4}{c}{LRW} & \multicolumn{4}{c}{VoxCeleb2}  \\
& {LMD$\downarrow$} & {CSIM$\uparrow$}  & {SSIM$\uparrow$}& {FID$\downarrow$}  & {LMD$\downarrow$} & {CSIM$\uparrow$} & {SSIM$\uparrow$} & {FID$\downarrow$}   \\

\hline
 {Chung et al.~\cite{chung2017you}} &  3.15 & 0.44 & \textbf{0.91}  & 132  &  5.4  & 0.29   &  0.79& 159   \\ 

 {Chen et al.~\cite{chen2019hierarchical}}  & 3.27   & 0.38  & 0.84  & 151 & 4.9  &0.31 &0.82 & 142  \\    
  {Vougioukas et al.~\cite{vougioukas2019realistic}}  & 3.62 & 0.35   & 0.88  &116 & 6.3 & 0.33  &\textbf{0.85}  & 127    \\   

  { Ours (K=1)} &  \textbf{3.13} &\textbf{ 0.49}  & 0.76 & \textbf{62} &\textbf{ 3.37}   & \textbf{0.42}  &  0.74  &\textbf{    47 }\\ 
  \end{tabular*}
    \label{tab:audio_tb}
\end{table}

\begin{table}[t]
     \caption{Quantitative results of different landmark to video methods on LRS3-TED and VoxCeleb2 datasets. Our model mentioned in this table are trained from scratch. We bold each leading score. The number after each method denotes the $K$ frames of the target subject.}
    \centering
    \begin{center}
    \begin{tabular*}{0.83\linewidth}{  l | c c c |  c c c }
      \toprule
      \hline
 Method & \multicolumn{6}{c}{Landmark-driven}   \\ \hline
Dataset & \multicolumn{3}{c}{LRS3-TED} & \multicolumn{3}{c}{VoxCeleb2}  \\
&{{{SSIM}}$\uparrow$}& {{{CSIM}}$\uparrow$} & 
{{{FID}}$\downarrow$}  &{{{SSIM}}$\uparrow$}  & {{{CSIM}}$\uparrow$}&{{{FID}}$\downarrow$}   \\
  
{Wiles et al.}~\cite{wiles2018x2face} & 0.57 &   0.28  &   {172}  &  0.65 & 0.31 &   117.3  \\ 
{{{Chen et al.}~\cite{chen2019hierarchical}}}& 0.66  & 0.31 & { 294 }   & 0.61  & 0.39 &  107.2    \\   
 {{{Zakharov et al.~\cite{zakharov2019few} (K=1)}}} & {  0.56  } &{     0.27    } &{  361    } &  0.64  &  0.42 &   88.0       \\ 
  {{{Wang et al.~\cite{wang2018fewshotvid2vid} (K=1)}}} & { 0.59} &{0.33 } &\textbf{{ 161}  }& 0.69 & \textbf{0.48} & {59.4}\\ 
 {{{Ours (K=1)} }}&\bf{ 0.71  } & \bf{  0.37   } &{  354   } &  \textbf{0.71}&0.44 &  \bf{ 40.8     } \\ 
 \hline
 {{{Zakharov et al.~\cite{zakharov2019few} (K=8)}}} & {    0.65  } &{   0.32   } &{  284 }  &0.71 &\textbf{ 0.54}  &  62.6     \\ 
  {{{Wang et al.~\cite{wang2018fewshotvid2vid} (K=8)}}} & {0.68 } &{0.36} & \textbf{144} &\textbf{0.72} & 0.53& {42.6 }  \\ 
 {{{Ours(K=8)} }}&\bf{ 0.76  } & \bf{   0.41  } &{  {324}   } &   0.71  & 0.50& \bf{   37.1   }  \\
 \hline
 Zakharov et al.~\cite{zakharov2019few}(K=32) & {  0.72    } &{    0.41   } & 154    & 0.75 & 0.54 &   51.8       \\ 
  {{{Wang et al.~\cite{wang2018fewshotvid2vid}(K=32)}}} & { 0.69} &{ 0.40} &{132 }  &0.73 & \textbf{0.60 } & {41.4}   \\ 
 {Ours (K=32)}&\bf{  0.79 } & \bf{   0.44  } &{ \textbf{122}    }& \bf{   0.78      } &  0.58& \textbf{ 35.7}   \\ 
  \end{tabular*}
  \end{center}
    \label{tab:lmark_tb}
\end{table}

 \begin{figure}
\includegraphics[width= 0.98 \linewidth]{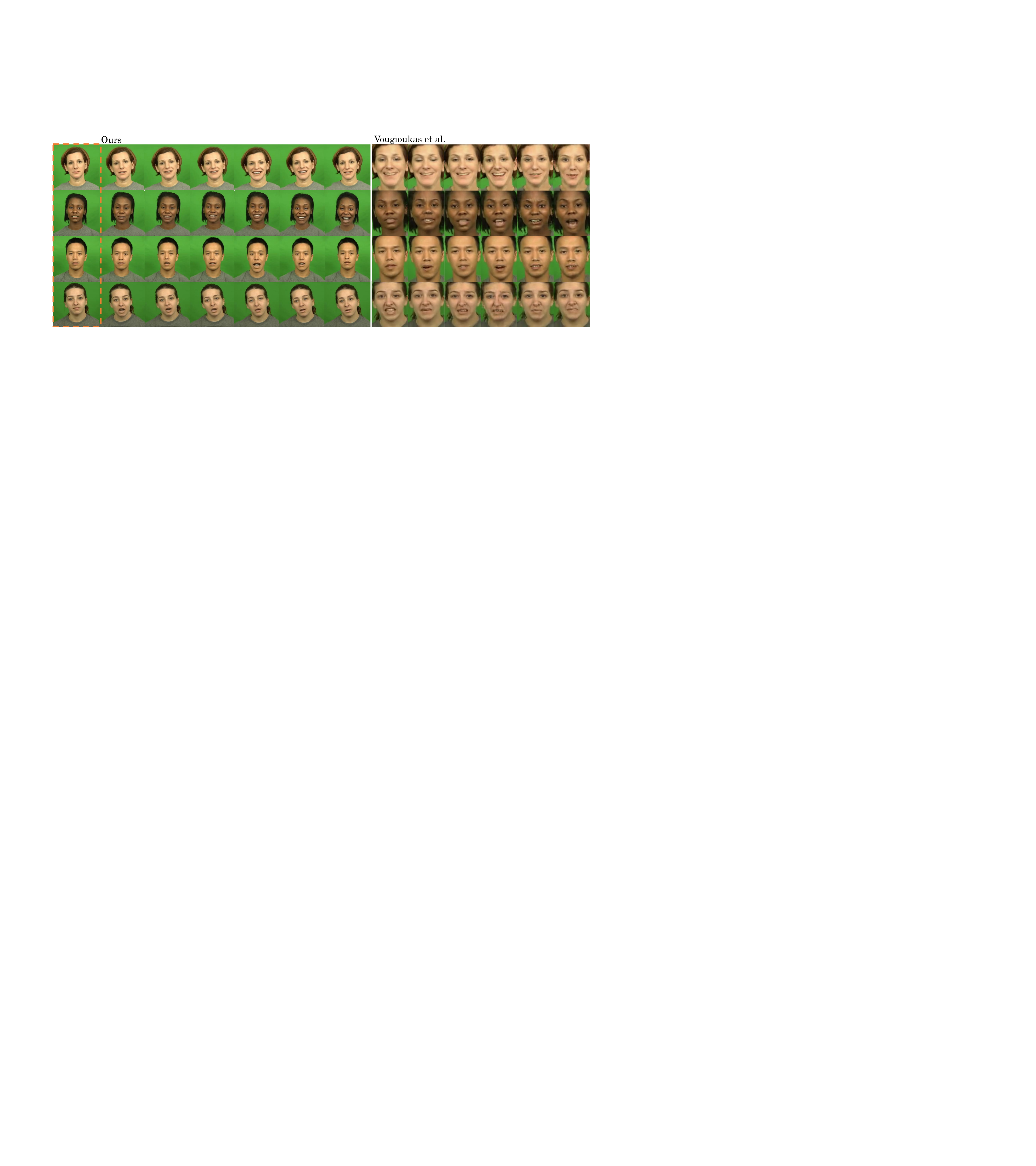}
\caption{Comparison of our model with Vougioukas et al.~\cite{vougioukas2019realistic} on CREMA-D testing set. The orange dashed box shows reference images. Our model is pretrained on VoxCeleb2 training set and then finetuned on CREMA-D.}
\label{fig:crema}
\end{figure}
\section{Results and Analysis}
\label{sec:quanti_res}
\noindent \textbf{Evaluation Metrics.} \quad We use several criteria for quantitative evaluation. We use Fréchet Inception Distance (FID)~\cite{heusel2017gans}, mostly quantifying the fidelity of synthesized images, structured similarity (SSIM)~\cite{wang2004image}, commonly used to measure the low-level similarity between the real images and generated images. To evaluate the identity preserving ability, same as \cite{zakharov2019few}, we use CSIM, which computes the cosine similarity between embedding vectors of the state-of-the-art face recognition network~\cite{deng2019arcface} for measuring identity mismatch. To evaluate whether the synthesized video contains accurate lip movements that correspond to the input condition, we adopt the evaluation matrix Landmarks Distance (LMD) proposed in~\cite{chen2018lip}. Additionally, we conduct user study to investigate the visual quality of the generated videos including lip-sync performance.
\begin{figure}
\includegraphics[width= \linewidth]{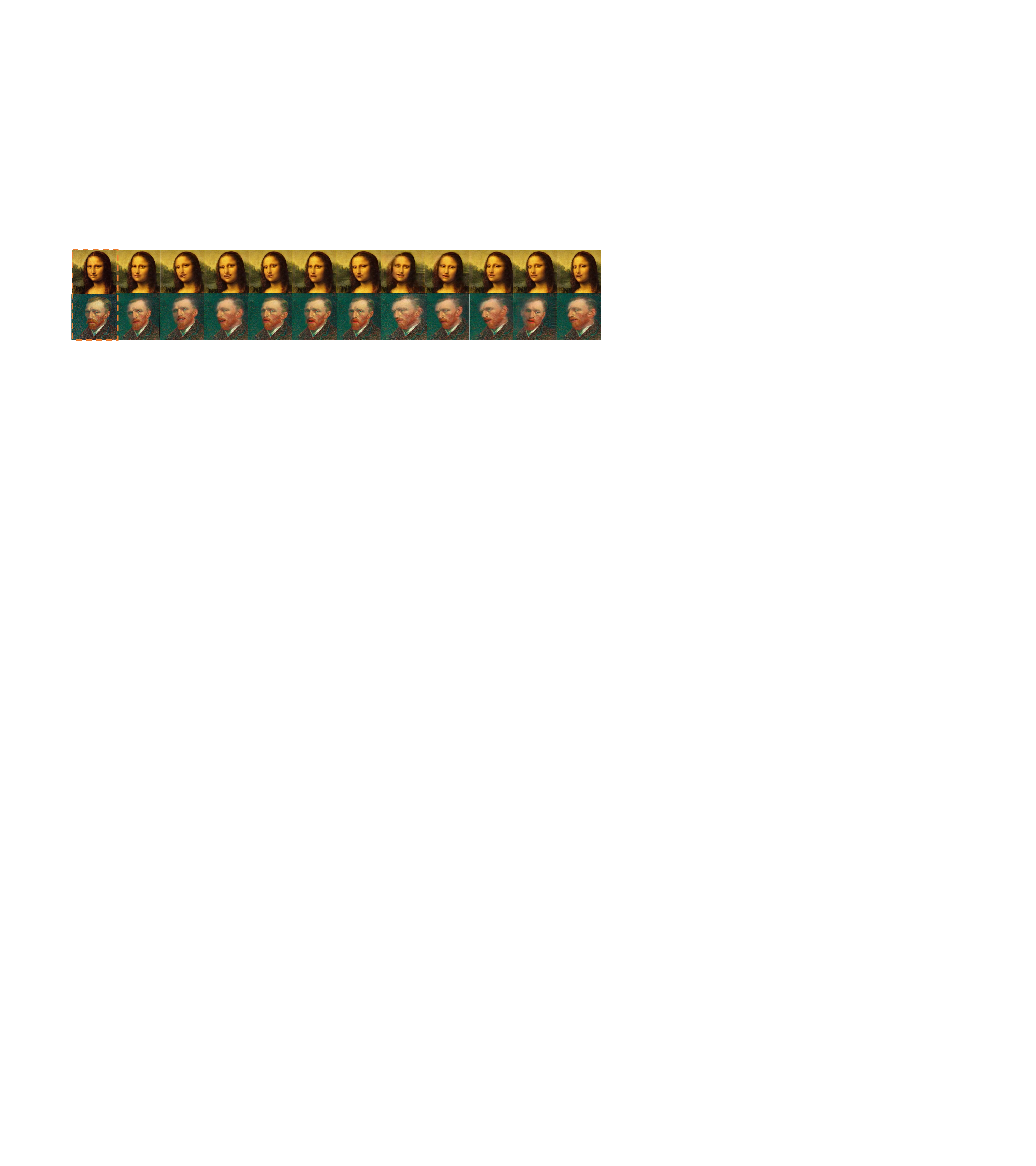}
\caption{We generate video frames on two real world images. The facial expressions are controlled by audio signal and the head motion is controlled by our head motion learner. The orange dashed box indicate reference frames.}
\label{fig:cartoon}
\end{figure}

\noindent \textbf{Comparison with Audio-driven Methods.} \quad We first consider such a scenario that takes audio and one frame as inputs and synthesizes a talking head saying the speech, which has been explored in Chung et al.~\cite{chung2017you}, Wiles et al.~\cite{wiles2018x2face}, Chen et al.~\cite{chen2019hierarchical}, Song et al.~\cite{ijcai2019-129}, Vougioukas et al.~\cite{vougioukas2019realistic}, and Zhou et al.~\cite{zhou2019talking}. Comparing with their methods, our method is able to generate vivid videos including controllable head movements and natural facial expressions. We show the videos driven by audio in the supplemental materials. Tab.~\ref{tab:audio_tb} shows the quantitative results. For a fair comparison, we only input one reference frame. Note that other methods require pre-processing including affine transformation and cropping under a fixed alignment. And we do not add such constrains, which leads to a lower SSIM matrix value (e.g., complex backgrounds). We also compare the facial expression generation (see Fig.~\ref{fig:crema})\footnote{We show the raw output from the network in Fig.~\ref{fig:crema}. Due to intrinsic limitations, \cite{vougioukas2019realistic} only generates facial regions with a fixed scale.} from audio with Vougioukas et al.~\cite{vougioukas2019realistic} on CREMA-D dataset~\cite{cao2014crema}. Furthermore, we show two generation examples in Fig.~\ref{fig:cartoon} driven by audio input, which demonstrates the generalizability of our network.

\begin{figure}
\includegraphics[width= \linewidth]{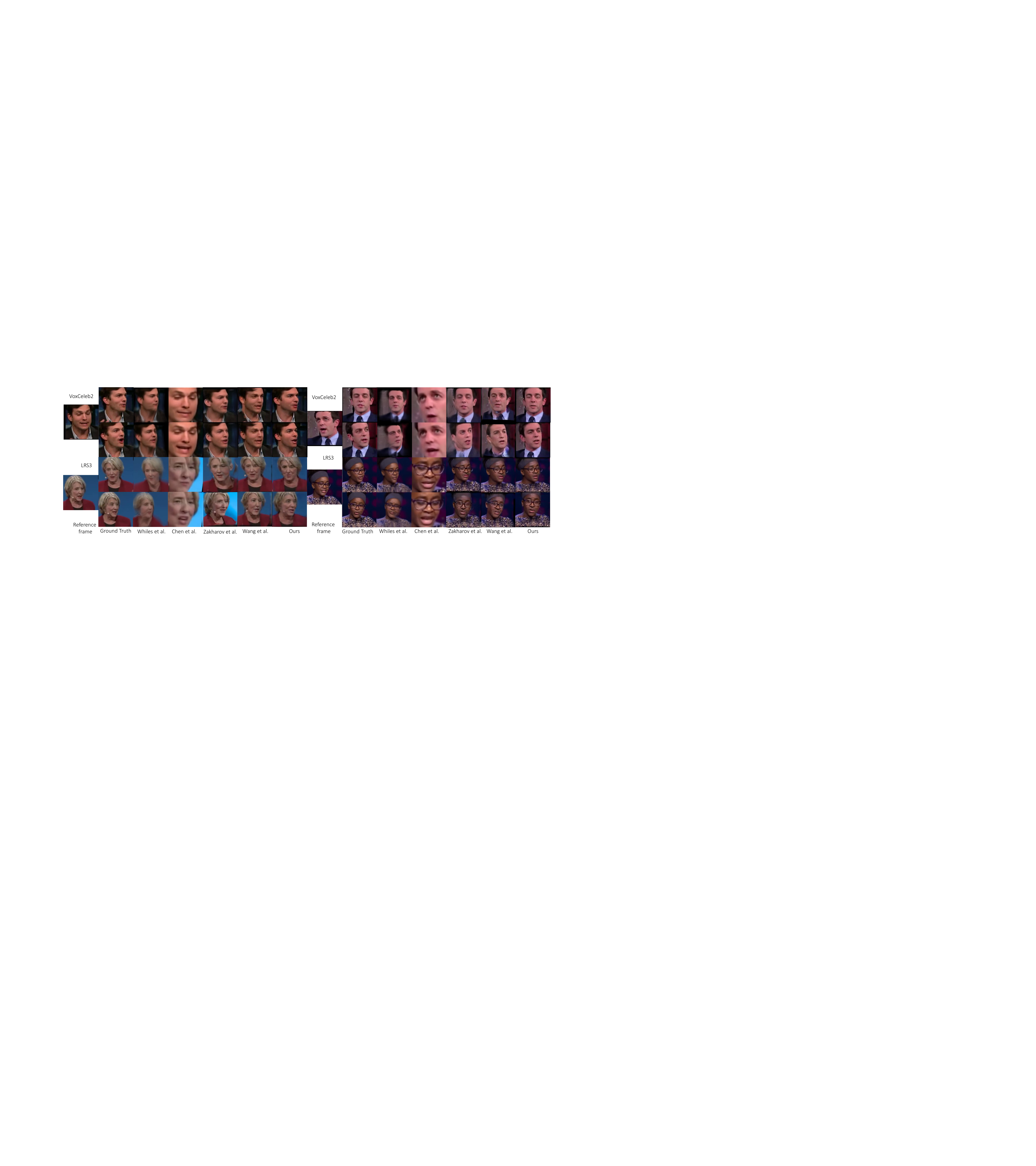}
\centering
\caption{Comparison on LRS3-TED and VoxCeleb2 testing set. We can find that our method can generate lip-synced video frames while preserving the identity information.} 
\label{fig:lmark_driven}
\end{figure}

\noindent \textbf{Comparison with Visual-driven Methods.} \quad We also compare our approach with state-of-the-art landmark/image-driven generation methods: Zakharov et al.\cite{zakharov2019few}, Wiles et al.~\cite{wiles2018x2face}, and Wang et al.~\cite{wang2018fewshotvid2vid} on VoxCeleb2 and LRS3-TED datasets. Fig.~\ref{fig:lmark_driven} shows the visual comparison, from where we can find that our methods can synthesize accurate lip movement while moving the head accurately. Comparing with other methods, ours performs much better especially when there is a apparent deformation between the target image and reference image. We attribute it to the 3D-aware module, which can guide the network with rough geometry information. We also show the quantitative results in Tab.~\ref{tab:lmark_tb}, which shows that our approach achieves best performance in most of the evaluation matrices. It is worth to mention that our method outperforms other methods in terms of the CSIM score, which demonstrates the generalizability of our method. We choose different $K$ here to better understand our matching scheme. We can see that with larger $K$ value, the inception performance (FID and SSIM) can be improved. Note that we only select $K$ images from the first 64 frames, and Zakharov et al.~\cite{zakharov2019few} and Wang et al.~\cite{wang2018fewshotvid2vid} select $K$ images from the whole video sequence, which arguably gives other methods an advantage.

\begin{figure}
\includegraphics[width= \linewidth]{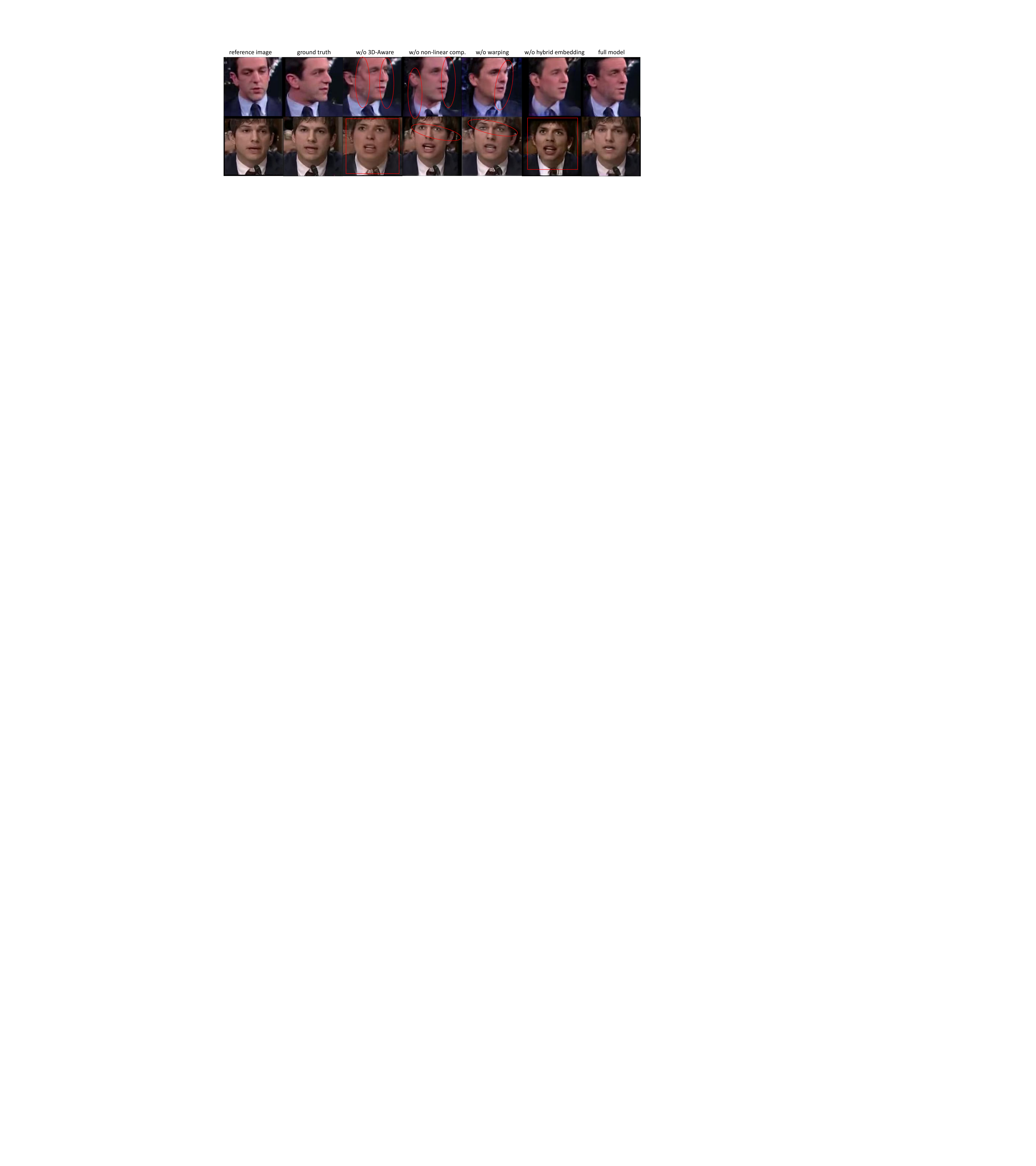}
\caption{Visualization of ablation studies. We show the artifacts using red circles/boxes. }
\label{fig:ablation}
\end{figure}
\noindent \textbf{Ablation Studies.} \quad We conduct thoughtful ablation experiments to study the contributions of each module we introduced in Sec.~\ref{sec:method} and Sec.~\ref{sec:3dgan}. The ablation studies are conducted on VoxCeleb2 dataset. As shown in Fig.~\ref{fig:ablation}, each component contributes to the full model. For instance, from the third column, we can find that the identity preserving ability drops dramatically if we remove the 3d-aware module. We attribute this to the valuable highly-aligned facial texture information provided 3D-aware module, which could stabilize the GAN training and lead to a faster convergence. Another interesting case is that if we remove our non-linear composition block or warping operation, the synthesized images may contain artifacts in the area near face edges or eyes. We attribute to the alignment issue caused by head movements. Please refer to supplemental materials for more quantitative results on ablation studies.

\begin{wrapfigure}{R}{0.3\textwidth}
\centering
\includegraphics[width=0.3\textwidth]{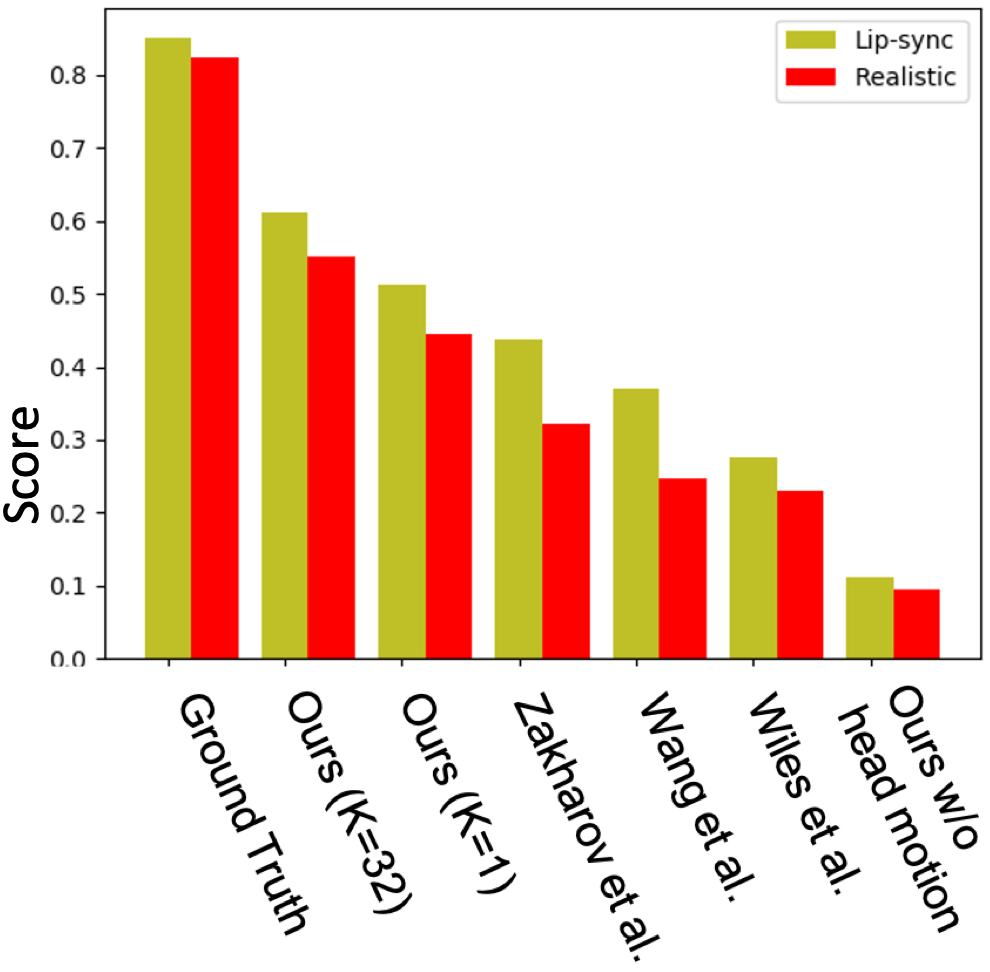}
\caption{The statistics of user studies. The scores are normalized to 1.}
\label{fig:user}
\end{wrapfigure}

\noindent \textbf{User Studies.} \quad 
To compare the photo-realism and faithfulness of the translation outputs, we perform human evaluation on videos generated by both audio and landmarks on videos randomly sampled from different testing sets, including LRW~\cite{Chung16}, VoxCeleb2~\cite{Chung18b}, and LRS3-TED~\cite{Afouras18d}. Our results including: synthesized videos conditioned on audio input, videos conditioned on landmark guidance, and the generated videos that facial expressions are controlled by audio signal and the head movements are controlled by our motion learner. Human subjects evaluation is conducted to investigate the visual qualities of our generated results compared with Wiles et al.~\cite{wiles2018x2face}, Wang et al.~\cite{wang2018fewshotvid2vid}, Zakharov et al.~\cite{zakharov2019few} in terms of two criteria: whether
participants could regard the generated talking faces as realistic and whether the generated talking head temporally sync with the corresponding audio. The details of User Studies are described in the supplemental materials. From Fig.~\ref{fig:user}, we can find that all our methods outperform other two methods in terms of the extent of synchronization and authenticity. It is worth to mention that the videos synthesized with head movements (ours, K=1) learned by the motion learner achieve much better performance that the one without head motion, which indicate that the participants prefer videos with natural head movements more than videos with still face.

\section{Conclusion and Discussion}
In this paper, we propose a novel approach to model the head motion and facial expressions explicitly, which can synthesize talking-head videos with natural head movements. By leveraging a 3d aware module, a hybrid embedding module, and a non-linear composition module, the proposed method can synthesize photo-realistic talking-head videos.
 
 Although our approach outperforms previous methods, our model still fails in some situations. For example, our method struggles synthesizing extreme poses, especially there is no visual clues in the given reference frames. Furthermore, we omit modeling the camera motions, light conditions, and audio noise, which may affect our synthesizing performance.
 
\footnotesize{\noindent \textbf{Acknowledgement.} \quad This work was supported in part by NSF 1741472, 1813709, and 1909912. The article solely reflects the opinions and conclusions of its authors but not the funding agents.}

\clearpage

\bibliographystyle{splncs04}
\bibliography{review}
\newpage

\appendix
\begin{center}
\textbf{\large Supplemental Materials}
\end{center}
We introduce the network details in Sec.~\ref{sec:network}. In Sec.~\ref{sec:results}, we show more results in qualitative and quantitative perspective. Note that the actual results of other comparison methods could be better, since we replicate them by ourselves. We will update those results once the code is publicly available. 
\begin{figure}
\centering
\includegraphics[width=0.8 \linewidth]{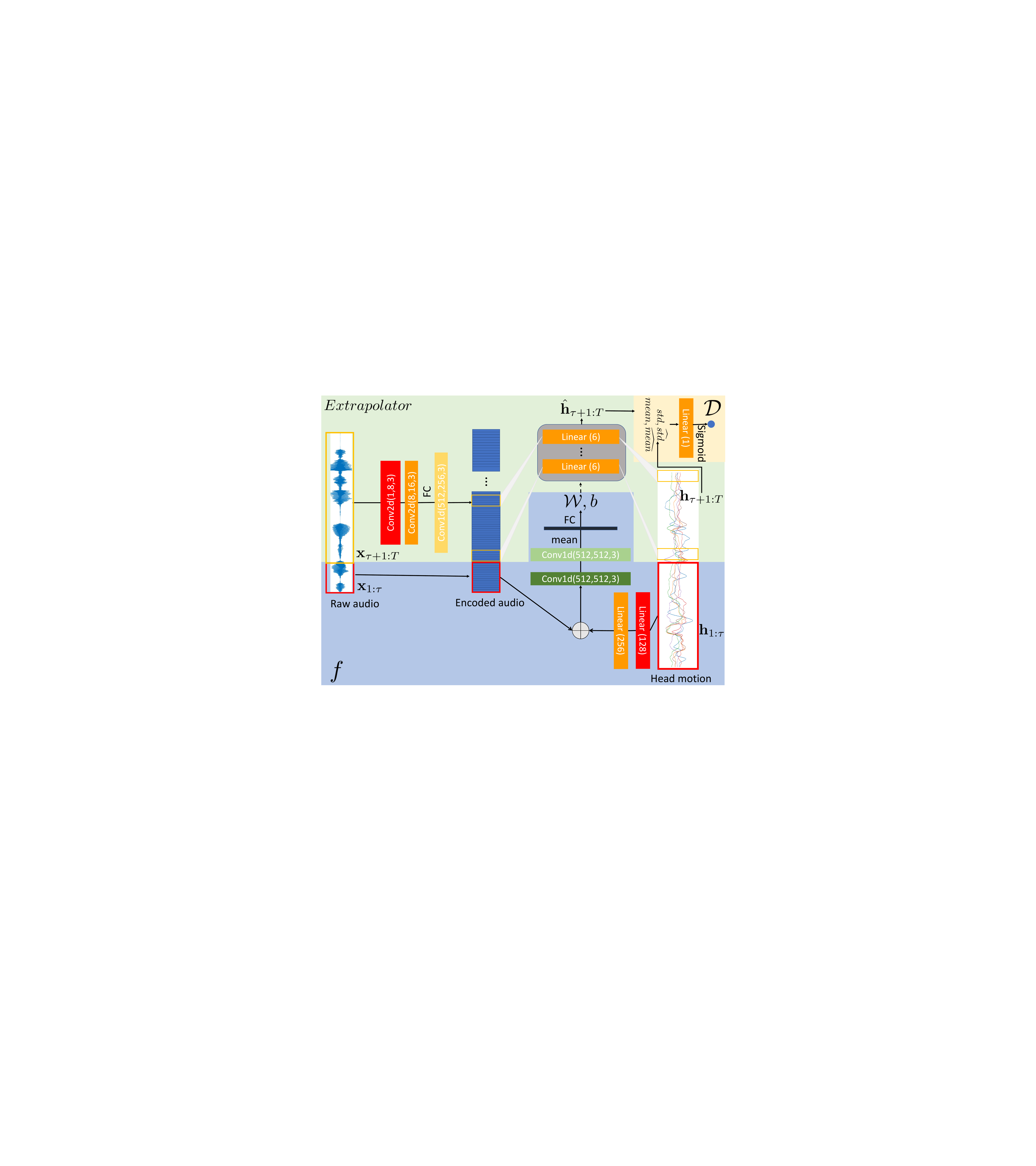}
\caption{The details of the head motion learner, which consists of a encoder $f$ (\textcolor{RoyalBlue}{blue} part), an extrapolator (\textcolor{OliveGreen}{green} part), and a discriminator (\textcolor{yellow}{yellow} part).}
\label{fig:motionlearner}
\end{figure}

\section{Network Details}
\label{sec:network}

\subsection{Details of The Head Motion Learner ($\Phi$)}
\label{subsec:motionlearner}

The head motion learner $\Phi$ consists of three sub-networks: a head motion encoding network $f$, a head motion extrapolation network $Extrapolator$, and a discriminator $\mathcal{D}$. Fig.~\ref{fig:motionlearner} shows the detailed network structure. Specifically, we first use $f$ to encode the raw audio $\mathbf{x}_{1:\tau}$ and its paired head motion $\mathbf{h}_{1:\tau}$ to network weights $\mathbf{w}$, which contains weights and biases $\{\mathcal{W},b\}$ for a linear layer. Since the audio sampling rate is $50000$ and the image sampling rate is $25$FPS, the size of the input $\mathbf{x}_{\tau+1:T}$ should be $(T-\tau)\times 0.04 \times 50000 $. At time step t, we use $\mathbf{x}_{t-3:t+4}$ to represent the audio signal. So, after stacking, the size of the input to $Extrapolator$ is $(7,(T-\tau)\times 0.04 \times 50000)$. In the $Extrapolator$, we apply two 2D convolutional layers on the inputs and then flatten it. Then we apply a 1D temporal convolution layer to encode it to audio feature with the size of $(T-\tau, 256)$. Then at each time step t, we forward
the feature chunk (size of $1,256$) to a linear layer, where the weights and biases $\{\mathcal{W},b\}$ are learned from $f$. Once the $Extrapolator$ generates all the fake head motion $\hat{\mathbf{h}}_{\tau+1:T}$, we forward it with ground truth motion ${\mathbf{h}}_{\tau+1:T}$ to discriminator $\mathcal{D}$. The $\mathcal{D}$ calculates the mean and standard deviation from real and fake sequences and then output a real/fake score based on the mean and standard deviation.
\subsection{Details of The Facial Expression Learner ($\Psi$)}
\label{subsec:expressionlearner}
\begin{figure}[ht]
\centering
\includegraphics[width=0.95 \linewidth]{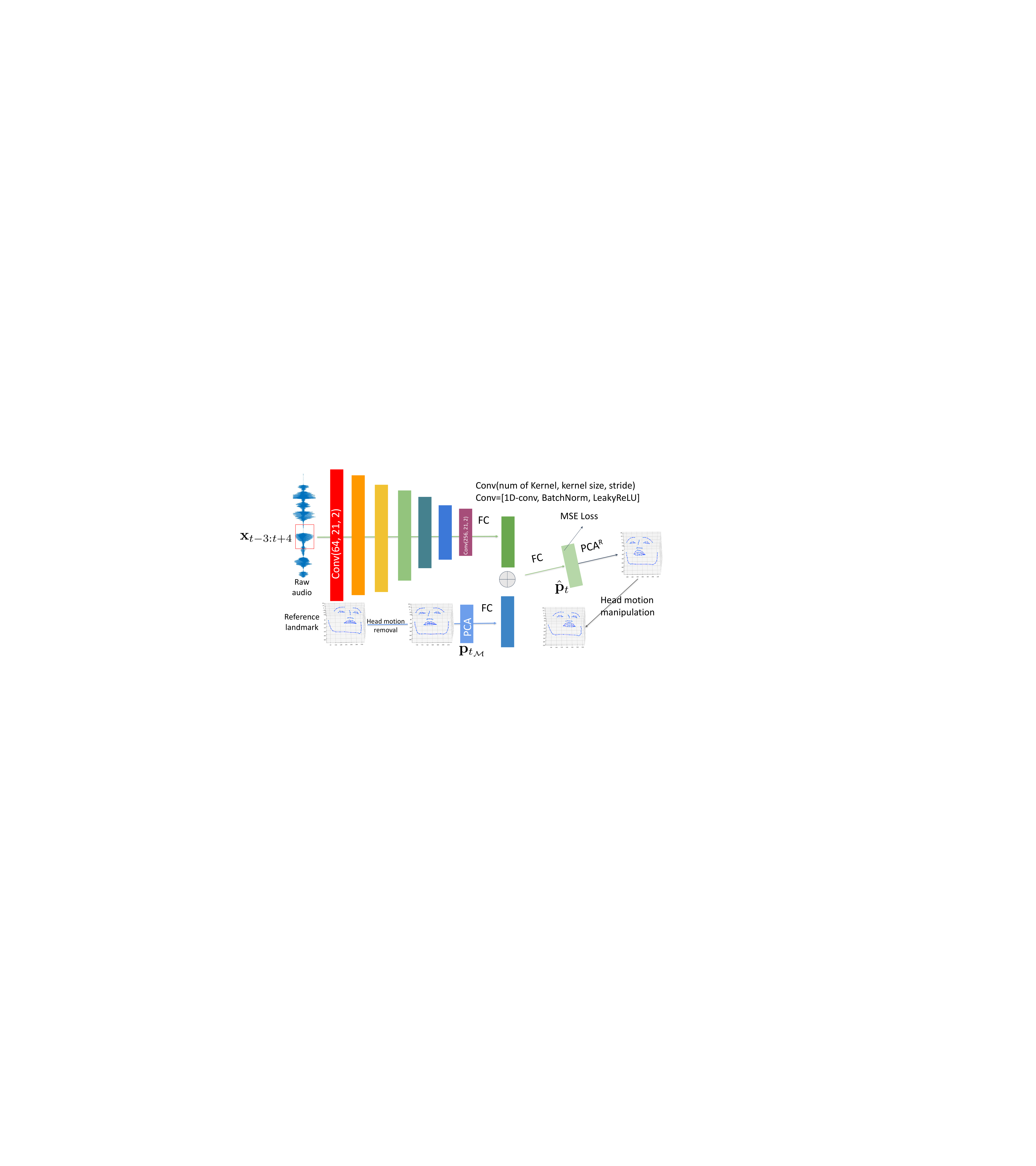}
\caption{The details of the facial expression learner. $\text{PCA}^{\text{R}}$ denotes reversed PCA operation.}
\label{fig:expressionlearner}
\end{figure}
The facial expression learner $\Psi$ is a linear regression network, which takes current audio chunk $\mathbf{x}_{t-3:t+4}$ and reference landmark PCA components $\mathbf{p}_{t_{\mathcal{M}}}$ as input and output current facial expression $\hat{\mathbf{p}}_{t}$. We use 20 PCA coefficients to represent facial expression. We list the details in Fig.~\ref{fig:expressionlearner}. We directly train and optimize the model on the PCA components output. During inference, we reconstruct the 3D landmark points from the 20 PCA coefficients.

\subsection{Details of The 3D Unprojection Network }
\label{subsec:unproject}
The unprojection network receives a RGB image $\mathbf{y}_{t_{\mathcal{M}}}$ and predict the position map image. We follow the same training  strategy and network structure as~\cite{feng2018joint}, which employs an encoder-decoder structure to learn the transfer function. we train the unprojection network on 300W-LP~\cite{zhu2016face}, since it contains face images across different angles with the annotation of estimated 3DMM coefficients, from which the 3D point cloud could be easily generated. We calculate MSE loss between the predicted position map and the ground truth position map. For training details, please refer to ~\cite{feng2018joint}.
\begin{figure}[t]
\centering
\includegraphics[width=0.95 \linewidth]{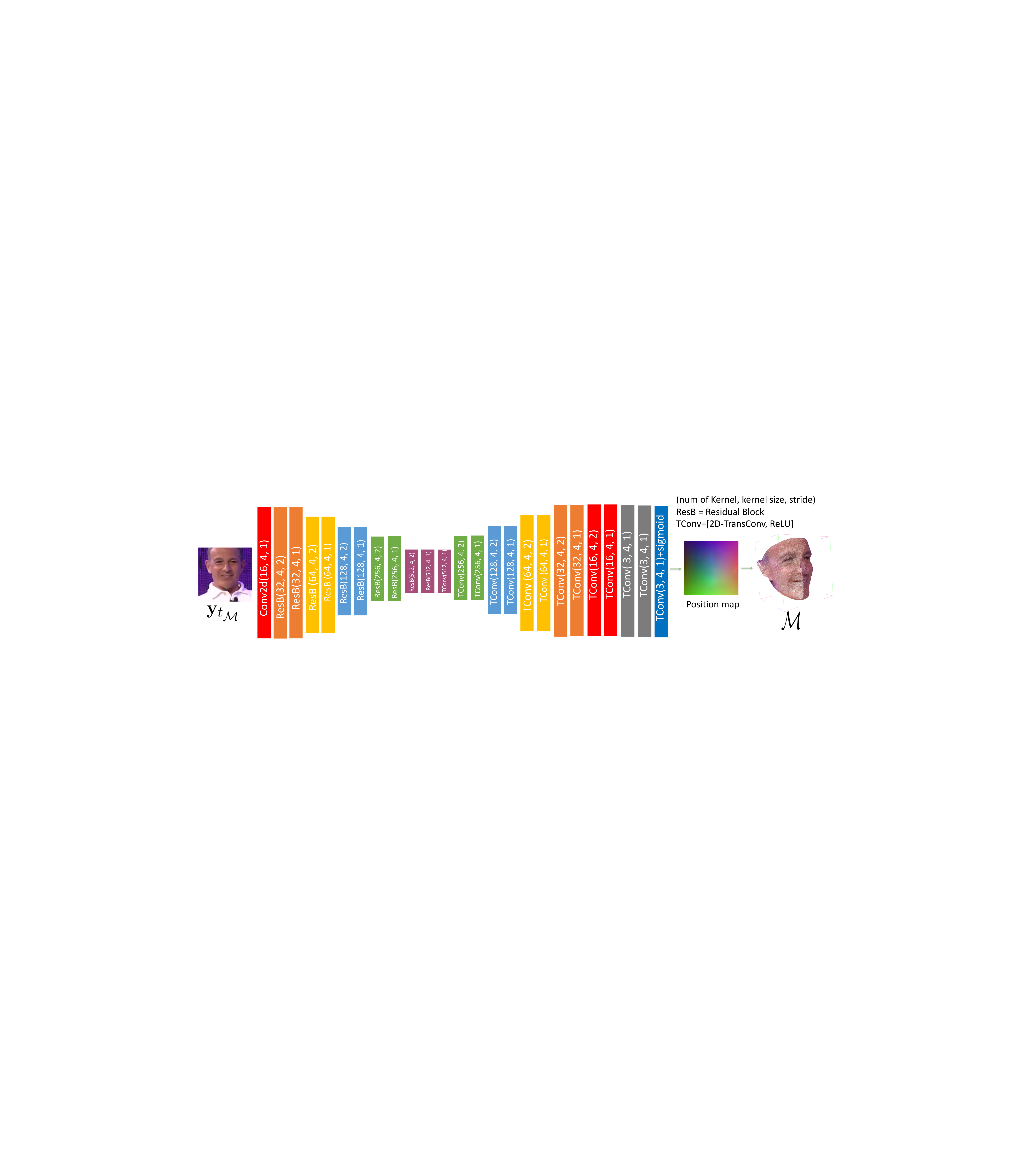}
\caption{The details of the unprojector. We use the method proposed in \cite{feng2018joint} as the unprojector.}
\label{fig:unprojector}
\end{figure}

\begin{figure}[ht]
\centering
\includegraphics[width=0.98 \linewidth]{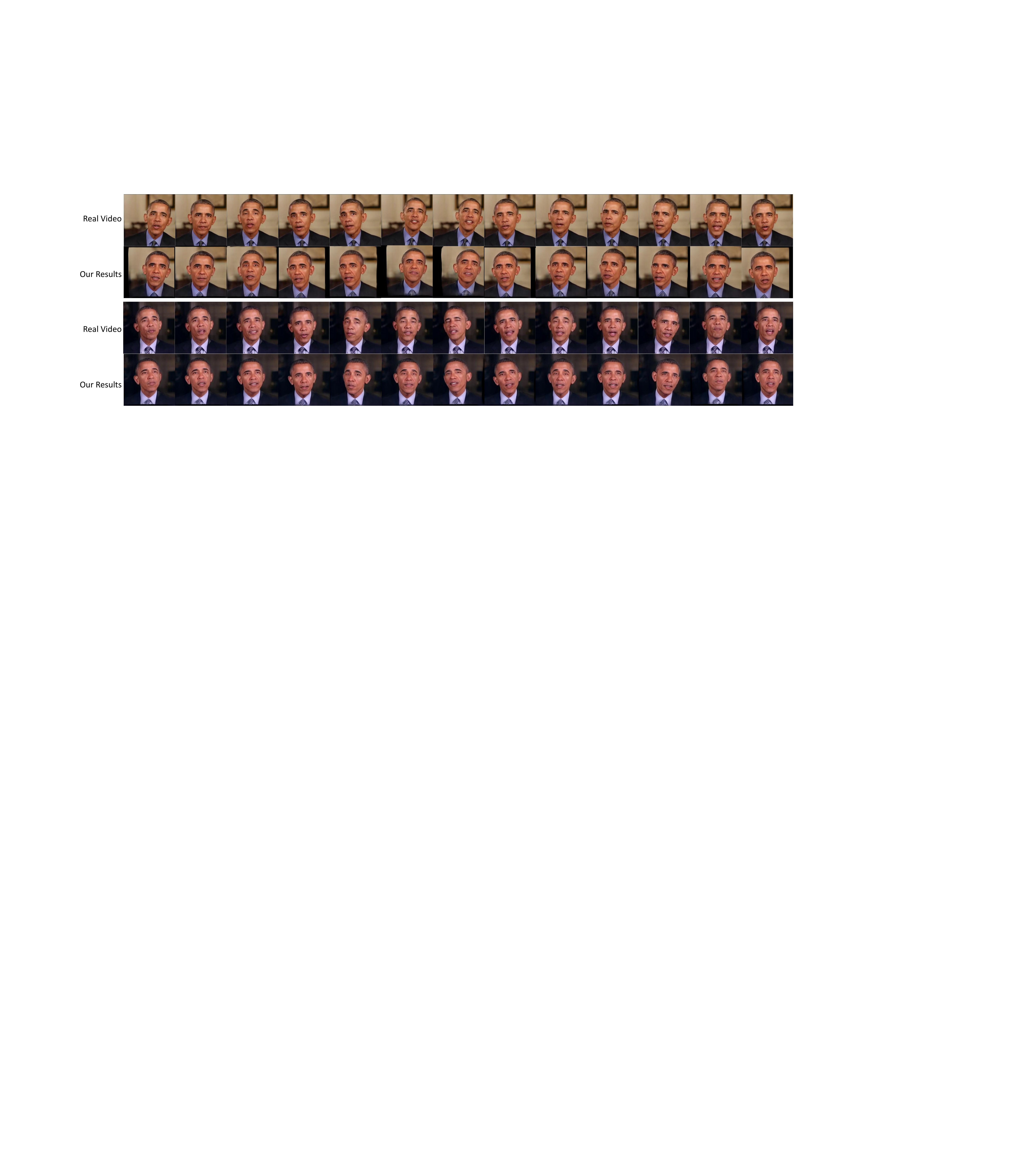}
\caption{The testing results on President Barack Obama's weekly address footage dataset ~\cite{suwajanakorn2017synthesizing}.}
\label{fig:obama}
\end{figure}

\section{More Results}
\label{sec:results}

\subsection{Controllable Videos}
\label{subsec:manipulation}
\label{sec:controllable}

\begin{figure}[ht]
\centering
\includegraphics[width=0.95 \linewidth]{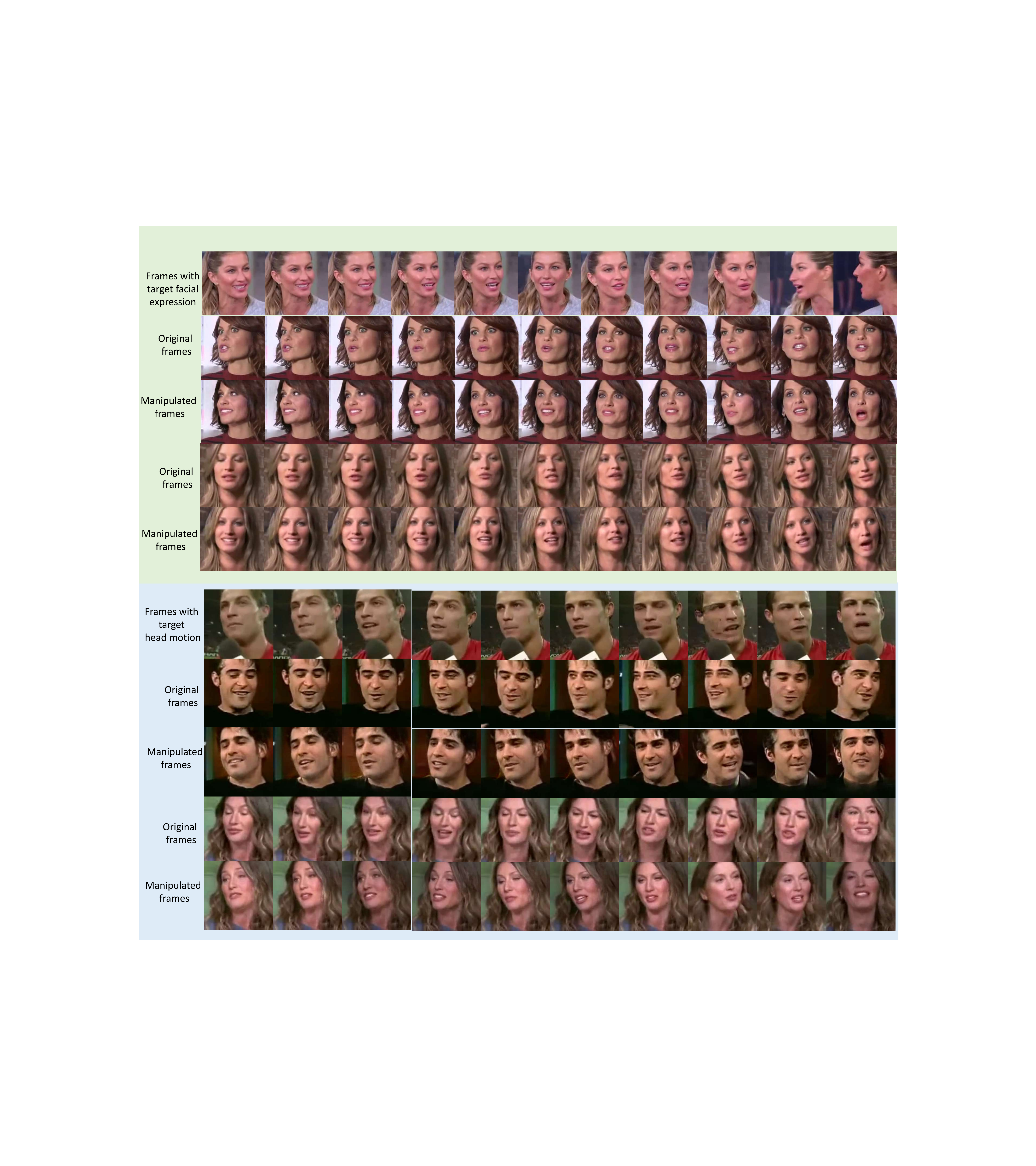}
\caption{The controllable results. The videos in upper part are manipulated with target facial expressions while keep the head motion unchanged. We show the target facial expression in the first row. The videos in lower part are manipulated with target head motion while keep the facial expression unchanged. }
\label{fig:manipularion}
\end{figure}

We show one testing results on VoxCeleb2 dataset to demonstrate our ability of generating controllable head motion and facial expression. From Fig.~\ref{fig:manipularion}, we can find that our model can generate controllable videos with desired head motion and facial expressions.

\subsection{Test on President Barack Obama Footage Dataset }
\label{subsec:fewshot}

To further improve the image quality, we fine-tuned our model with $K=8$ on five videos from the President Barack Obama's weekly address footage dataset ~\cite{suwajanakorn2017synthesizing} and leave the rest videos as testing set. Fig.~\ref{fig:obama} shows two example testing results.
\begin{table}
  \caption{Ablation studies on VoxCeleb2 dataset. Our model mentioned in this table are trained from scratch.}
    \centering
  \begin{tabular*}{0.54\linewidth}{  l |  c c c  }
      \toprule
      \hline
Method & {CSIM$\uparrow$} &  {SSIM$\uparrow$}&{FID$\downarrow$}   \\ \hline
{Baseline} & 0.19 &0.67 & 112\\ 
{Full Model} & 0.44 &0.71  & 40.8\\ 
{w/o 3D-Aware} &0.21  &0.61  & 109\\ 
{w/o Hybrid-Attention} & 0.37 & 0. 73 &{57.8}  \\ 
{w/o Non-Linear Comp.}  &  0.40 &{ 0.69} & 64.5  \\ 
{w/o warping}  &  { 0.34} &{ 0.67} & 78.2  \\ \hline
  \end{tabular*}
    \label{tab:alation}
\end{table}
\subsection{Ablation Studies}
\label{subsec:ablation}
We conduct ablation experiments to study the contribution of four components: 3D-Aware, Hybrid-Attention, Non-Linear Composition, and warping. The Baseline model is a straightforward model without any features (e.g. 3D-Aware, Hybrid-Attention). Table.~\ref{tab:alation} shows the quantitative results of ablation studies.

\subsection{Settings of User Studies}
\label{subsec:userstudy}

Human subjects evaluation is conducted to investigate the visual qualities of our generated results compared with Zakharov et al.~\cite{zakharov2019few}, Wang et al.~\cite{wang2018fewshotvid2vid} and Wiles~\cite{wiles2018x2face}. The ground truth videos are selected from different sources: we randomly select samples from the testing set of LRW~\cite{Chung16}, VoxCeleb2~\cite{Chung18b} and LRS3~\cite{Afouras18d}. Three methods are evaluated w.r.t. two different criteria: whether participants could regard the generated videos as realistic and whether the generated talking-head videos temporally sync with the corresponding audio. We shuffle all the sample videos and the participants are not aware of the mapping between videos to methods. They are asked to score the videos on a scale of 0 (worst) to 10 (best). There are overall 20 participants involved (at least $50\%$ of them are native English speaker), and the results are averaged over persons and videos.

\end{document}